\documentclass[journal]{IEEEtran}
\usepackage{graphicx}
\usepackage{amsmath}
\usepackage{amssymb}
\usepackage{multirow}
\usepackage[dvipsnames, svgnames, x11names]{xcolor}
\usepackage{graphicx,times}
\usepackage{subfigure}         
\ifCLASSINFOpdf
\else
\fi
\newcommand{\squishlist}{
 \begin{list}{$\bullet$}
  { \setlength{\itemsep}{0pt}
     \setlength{\parsep}{3pt}
     \setlength{\topsep}{3pt}
     \setlength{\partopsep}{0pt}
     \setlength{\leftmargin}{1.5em}
     \setlength{\labelwidth}{1em}
     \setlength{\labelsep}{0.5em}}
     }

\newcommand{\squishlisttwo}{
 \begin{list}{$\bullet$}
  { \setlength{\itemsep}{0pt}
     \setlength{\parsep}{0pt}
    \setlength{\topsep}{0pt}
    \setlength{\partopsep}{0pt}
    \setlength{\leftmargin}{2em}
    \setlength{\labelwidth}{1.5em}
    \setlength{\labelsep}{0.5em} } }

\newcommand{\squishend}{
\end{list} }

\begin{document}
	\title{MHSA-Net: Multi-Head Self-Attention Network for Occluded Person Re-Identification }

	\author{Hongchen Tan,   Xiuping Liu,   Baocai Yin  and Xin Li,~\IEEEmembership{Senior Member, IEEE}

		\thanks{(Corresponding author: Baocai Yin)
			
			Hongchen Tan and  Baocai Yin are  with Artificial Intelligence Research Institute, Beijing University of Technology, Beijing 100124, China (e-mail: tanhongchenphd@bjut.edu.cn; ybc@bjut.edu.cn).

			Xin Li is with School of Electrical Engineering \& Computer Science, and Center for Computation \& Technology, Louisiana State University, Baton Rouge (LA) 70808, United States of America (e-mail: xinli@cct.lsu.edu).
			
			Xiuping Liu is with School of Mathematical Sciences, Dalian University of Technology, Dalian 116024, China (xpliu@dlut.edu.cn.).

		}
		\thanks{}
		\thanks{}}

	\markboth{Journal of \LaTeX\ Class Files,~Vol.~14, No.~8, August~2015}%
	{Shell \MakeLowercase{\textit{et al.}}: Bare Demo of IEEEtran.cls for IEEE Journals}
	%
	
	\maketitle
\begin{abstract}
	
This paper presents a novel person re-identification model, named Multi-Head Self-Attention Network (MHSA-Net), to prune unimportant information and capture key local information from person images. 
MHSA-Net contains two main novel components: Multi-Head Self-Attention Branch (MHSAB) and Attention Competition Mechanism (ACM). 
The MHSAB adaptively captures key local person information, and then produces effective diversity embeddings of an image for the person matching. 
The ACM further helps filter out attention noise and non-key information. 
Through extensive ablation studies, we verified that the Multi-Head Self-Attention Branch (MHSAB) and Attention Competition Mechanism (ACM) both contribute to the performance improvement of the MHSA-Net. Our MHSA-Net achieves  competitive  performance  in the  standard  and occluded person Re-ID tasks. 

\end{abstract}

\begin{IEEEkeywords}
Occluded Person Re-ID, Multi-Head Self-Attention, Attention Competition Mechanism, Feature Fusion.
\end{IEEEkeywords}

\IEEEpeerreviewmaketitle
\section{Introduction}~\label{Introduction}

Person re-identification (Re-ID) is a fundamental task in distributed multi-camera surveillance. It identifies the same person in different (non-overlapping) camera views. 
Re-ID has important applications in video surveillance and criminal investigation.  
With the surge of interest in deep representation learning, the person Re-ID task  has
achieved  great progress in recent years~\cite{9336268}.  
Although recently   many methods~\cite{ZhengWang2020,  8985292,  9397375,  9513260, Kaiyang19,  Zhun2020, WU2021107424}  have  boosted  the  performance  of the  standard person Re-ID  task, they didn't consider the situation that the person is occluded by various obstructions like cars, trees, or other people.  The   occlusion in person images  is  still  a key challenging issue that hinders Re-ID performance.  
Thus, this paper aims to develop a  Re-ID algorithm that can better handle occlusions in images.

\begin{figure}[h!tb]
\centering
	\begin{center} 
		\includegraphics[scale=0.22]{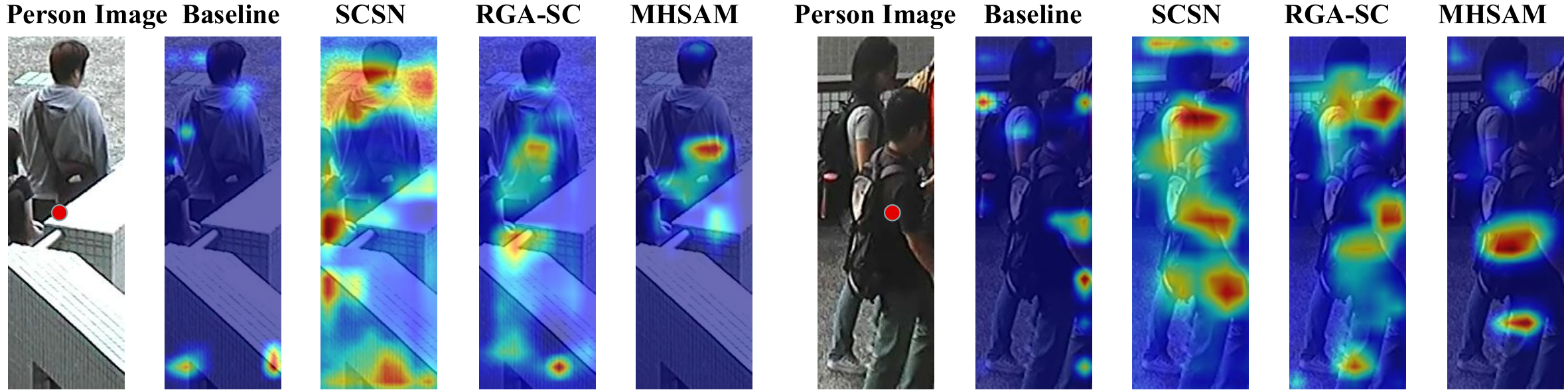}
	\end{center} 
	\caption{The occluded  person  images'  attention maps  are   produced  by  our  Baseline,  RGA-SC~\cite{9157488}, SCSN (3-stage)~\cite{9156982}   and  the  person  Re-ID  model  equipped  with   Multi-Head Self-Attention mechanism (MHSAM)~\cite{Ashish2017, 2020Jean-Baptiste}. The red dot is the target  person.  } 
	\label{Fig1} 
\end{figure}

In the occluded person Re-ID task, occluded regions often contain a lot of noise that results in mismatching.
So a key issue in occluded Re-ID is to build discriminative features from unoccluded regions.
Some part-based methods~\cite{WeiShi2015, Lingxiao2018, Lingxiao2018xl}  manually crop the occluded target person in probe images and then use the unoccluded parts as the new query. 
However, these manual operations are inefficient in practice. 
Another type of approach is to use human model to help build person features. More recently, \cite{Jiaxu112019, Guanan2020, Pose-guided2020} applied pose estimators to obtain the person's key points to locate effective regions of the person. 
However, the difference between training datasets of pose estimation and that of person retrieval often exist, making pose estimation based feature extraction sometimes unstable. 
It is desirable to design an effective mechanism to adaptively capture the key features from non-occlusion regions without relying on human models.

We are inspired by the recent Multi-Head Self-Attention mechanism (MHSAM)~\cite{Ashish2017, 2020Jean-Baptiste, DBLP01232, ZHAO2022360}, which flexibly captures spatially different local salience from the whole image, and generates multiple attention maps, from different aspects, for a single image.
With MHSAM, noisy/unimportant regions can be pruned and key local feature information can be highlighted. 
Therefore, we believe the idea of MHSAM can help a Re-ID model to better locate key features from occluded images. 
As shown in Fig.~\ref{Fig1}, compared with the Baseline,  two  outstanding  attention Re-ID model    RGA-SC~\cite{9157488}  and SCSN (3-stage)~\cite{9156982}, the MHSAM can help  the  person  Re-ID  model  better capture key information of the target person from the unoccluded regions and avoid information from occluded regions. 
The baseline may undesirably pay attention to clutter regions (left example) or other persons (right example), while our MHSAM model handles such occlusions much better.

However, developing effective MHSAM for the task of Re-ID is non-trivial and needs careful design. 
We propose a novel MHSAM module for the person Re-ID task with a set of new strategies to help select the key sub-regions in the image. 
We call our new attention module a \emph{Multi-Head Self-Attention Branch} (MHSAB). 

Furthermore, attention noise from the occluded or non-key regions often exists, and it affects the performance of the Re-ID model, we propose to design an \emph{Attention Competition Mechanism} (ACM) to further  help MHSAB suppress or filter out such attention noise from non-key sub-regions. 
Our main contributions are as follows:

\squishlist
\item[(I)] 
We proposed a new attention module, MHSAB, that can more effectively extract person features in occluded person Re-ID. 
\item[(II)]
We proposed a new attention competition module (ACM) to better prune attention noise from unimportant regions. 
\item[(III)] By integrating MHSAB and ACM modules, our final 
MHSA-Net framework demonstrates better performance over most state-of-the-art methods when processing occluded images, i.e., on four occlusion datasets: Occluded-DukeMTMC~\cite{Jiaxu112019},  P-DukeMTMC-reID~\cite{Jiaxuan2018}, Partial-REID~\cite{WeiShi2015}, and Partial-iLIDS~\cite{Lingxiao2018}.
On standard generic person Re-ID datasets, e.g., Market-1501~\cite{Zheng2015Scalable}, DukeMTMC-reID~\cite{Ristani2016Performance,ZhedongZheng}, and CUHK03~\cite{Weireid2014}, our MHSA-Net produces similar results with these state-of-the-art algorithms. 
\squishend

\section{Related  Work}

\subsection{Attention Mechanism in Person Re-identification}
Attention mechanisms have been widely exploited in computer vision and natural language processing, for instance in Text-to-Image Synthesis~\cite{thc2019}, Object Tracking~\cite{He2018A}, Image/Video Captioning~\cite{Lu2016Knowing}, Visual Question Answering~\cite{Zichao2016},  Neural Machine Translation~\cite{Dzmitry14},  and  some Video  Tasks~\cite{8718523, 8955791, 9399800}   
It can effectively capture task-relevant information and reduce interference from less important ones. 
Recently, many person Re-ID approaches~\cite{Yeong-Jun16, Kalayeh2018Human, Chunfeng, Xuelin18, Chi17, Jing181,  Tianlong2019, Bryan2019} also introduced various attention mechanisms into deep models to enhance identification  performance.

\cite{Kalayeh2018Human, Chunfeng, Xuelin18, Chi17} applied
a human part detector or a human parsing model to capture features of body parts. 
\cite{Dongreid2014} explored both the human part masks and human poses to enhance human body feature extraction. 
\cite{Jing181, Jiaxu112019} exploited the connectivity of the key points to generate human part masks and focuses on the human's representation.   However, the success of such approaches heavily relies on the accuracy of the human parsing models or pose estimators.

Other methods  typically focus  on   extracting  the person appearance or gait information,  from the 3D  space or depth images,   to reduce the interference of background or occlusion. For example, Zhedong et al  ~\cite{Zhedong3Dreid} try  to    project   the 2D person  image into the  3D  space, and  conduct  the  person  matching in  the 3D space.    Munaro et al.~\cite{Munaroreid} proposed point cloud matching  (PCM)  stragy to compute the distances of multi-view point cloud sets, so as to distinguish between different persons.  Haque et al. ~\cite{7780507} adopted  3D LSTM to build  motion dynamics of 3D person point clouds for person matching. 
~\cite{9466418} proposed a self-supervised gait encoding approach that can leverage unlabeled 3D skeleton data to learn gait representations for person Re-ID. 
Sivapalan et al.~\cite{6117504} extended  the Gait Energy Image (GEI) ~\cite{5479416} to 3D domain and proposed Gait Energy Volume (GEV) strategy  based on depth images to perform gait-based person  Re-ID.  In ~\cite{DBLPLCS17}, \textcolor{red}{Convolutional Neural Network Long Short-Term Memory (CNN-LSTM)} with reinforced temporal attention (RTA) was proposed for person matching  based on a split-rate RGB-Depth transfer method.

Besides, many  methods~\cite{Tianlong2019, Bryan2019, li2018harmonious, AANET19, Yifan2018} tried to exploit a different type of attention mechanism that does not need to use human models to capture human body features.  
\cite{Jianlou2018} proposed a dual attention matching network based on an inter-class and an intra-class attention module to capture context  information of video sequences for person Re-ID.  ABD-Net~\cite{Tianlong2019} combined spatial and channel attention to directly learn human's information from the data and context.  \cite{Yifan2018} calculated the similarity of the local features to enhance local part information. 
\cite{AANET19} applied an attribute classification to gain local attention information. 
However, it does not consider how to filter out information from the occlusion regions in the image. 
Therefore, with its fixed and parameter-free attention patterns, information from the occlusion region will be inevitably included.

Similar to \cite{Tianlong2019, Jianlou2018, AANET19, Yifan2018}, our attention module also does not rely on an external human model. 
But different from these methods, our attention mechanism can adaptively enhance/suppress attention weights of local features 
through a multi-parameter learning strategy. 
The attention information of occluded and unoccluded regions in our attention mechanism is adaptively adjusted according to the targeting task. 
For the person Re-ID task, our attention module can flexibly capture the key local features and prune out information from occlusion regions.

\begin{figure*}[h!tb]
	\centering
	\begin{center} 
		\includegraphics[scale=0.31]{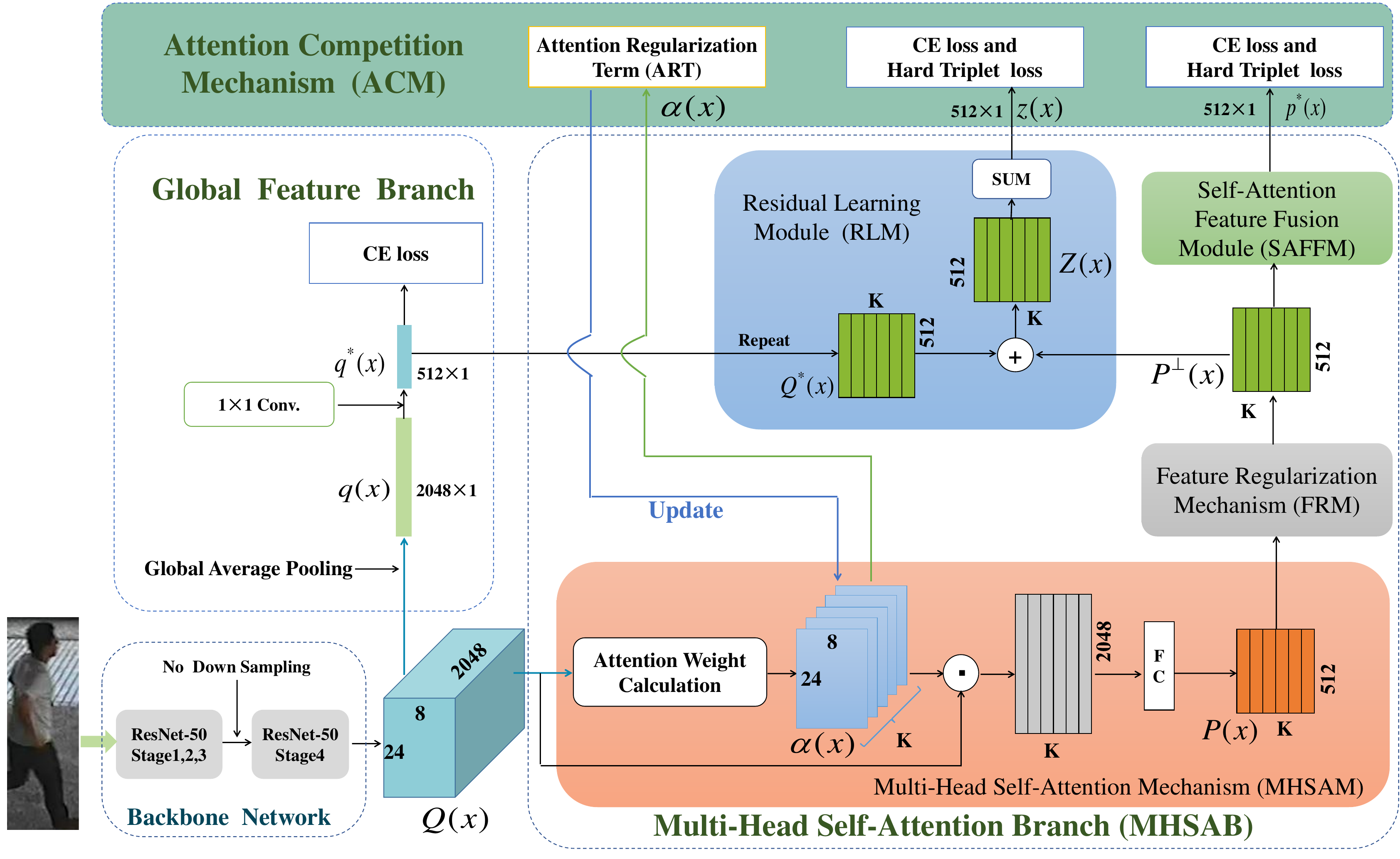}
	\end{center} 
	\caption{The architecture of Multi-Head Self-Attention  Network (MHSA-Net) for occlusion/standard person Re-ID task. The MHSA-Net contains three modules:  the Global Feature Branch, the Multi-Head Self-Attention Branch (MHSAB), and the Attention Competition Mechanism (ACM).  The CE loss denotes  the cross entropy  loss  function. }
	\label{Fig:Pipeline_MHSA} 
\end{figure*}

\subsection{Occluded Person Re-identification}

Occlusion is a key challenging issue in person Re-ID. 
Recent studies~\cite{WeiShi2015, Lingxiao2018, Lingxiao2018xl, Jiaxuan2018, Jiaxu112019, LingxiaoHE19, Pose-guided2020, Guanan2020} on this topic can be divided into two categories: (i) partial person Re-ID methods~\cite{WeiShi2015, Lingxiao2018, Lingxiao2018xl}, and (ii) occluded person Re-ID methods~\cite{Jiaxuan2018, Jiaxu112019, LingxiaoHE19, Pose-guided2020, Guanan2020}. 

The former category aims to match a partial probe image to a gallery holistic image. For example, \cite{WeiShi2015} adopted a
global-to-local matching mechanism to capture the key information from the spatial channel  of  the  feature maps.   DSR~\cite{Lingxiao2018, Lingxiao2018xl} proposed a spatial feature reconstruction strategy to align  the partial person image with holistic images.  However, these methods need a manual crop of the  occluded target person in the probe image, before the cropped unoccluded part can be used to retrieve the target person. 

The latter category aims to directly capture key features from the whole occluded person image to perform the person matching.   AFPB~\cite{Jiaxuan2018} combined the  occluded/unoccluded classification task and  person ID classification task to improve the performance of  deep  model on  capturing key information. 
FPR~\cite{LingxiaoHE19} reconstructed the feature map of unoccluded regions in occluded person  image and further improved it by a foreground-background mask to avoid the influence of background clutter.  \cite{Jiaxu112019, Pose-guided2020, Guanan2020}  proposed pose guided feature alignment methods to match the local patches of query and the gallery images based on human key-points. 
Our MHSA-Net also belongs to this type of method. However, different from these methods, the MHSA-Net does not require any additional model. Also, the MHSA-Net can more effectively capture unoccluded local information. 

\begin{figure*}
	\centering
	\begin{center} 
		\includegraphics[scale=0.3]{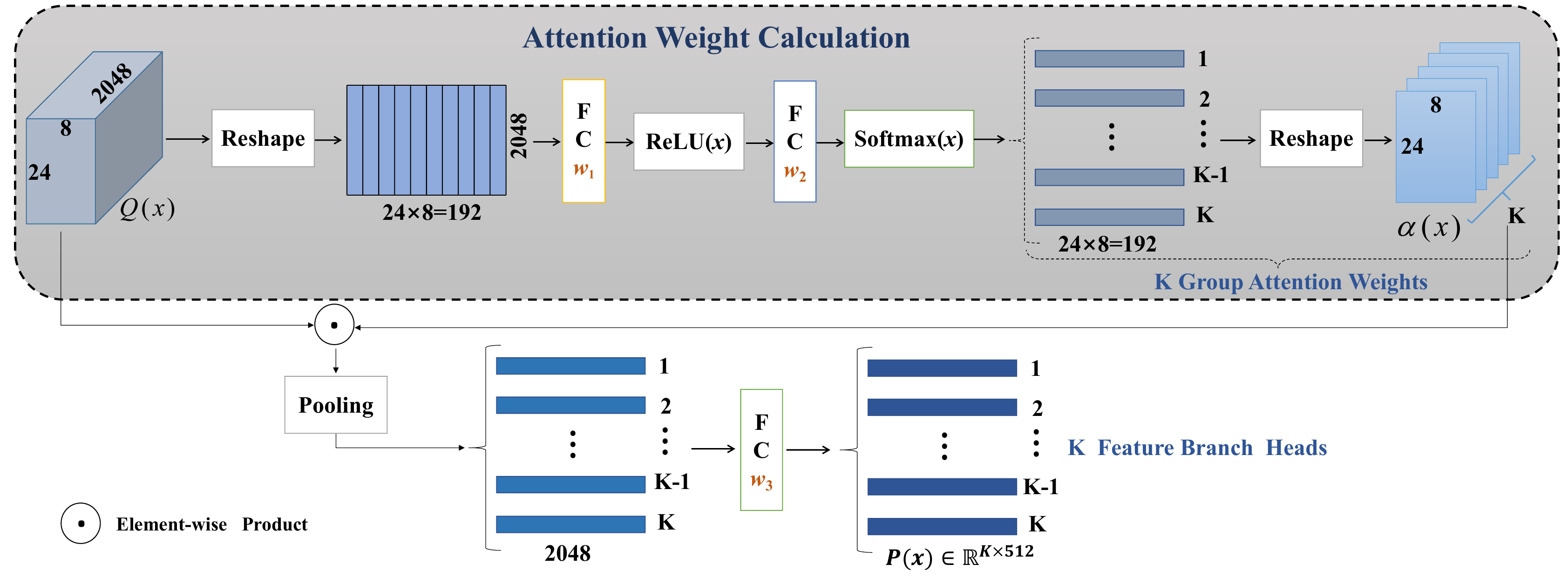}
	\end{center} 
	\caption{The  architecture    of  the  Multi-Head Self-Attention Mechanism (MHSAM).} 
	\label{Fig_MHSAM_1} 
\end{figure*}

\section{MHSA-Net overview}

Our MHSA-Net contains three modules:  the Global Feature Branch, the Multi-Head Self-Attention Branch (MHSAB), and the Attention Competition Mechanism (ACM), as illustrated in Fig.~\ref{Fig:Pipeline_MHSA}. 

The Global Feature Branch computes a basic large feature tensor  $Q(x)$ and a global feature $q^{*}(x)$ for MHSAB and person matching (CE loss). 
We use the widely adopted Backbone Network ResNet-50~\cite{KaimingRSE2015} to compute feature tensor $Q(x)$, then down-sample it to $q^{*}(x)$ for MHSAB and person matching. 

The MHSAB, the core component in MHSA-Net, captures the key local information and outputs the fusion feature $p^{*}(x)$ for the person matching. The MHSAB contains four sub-modules: the (1) Multi-Head Self-Attention Mechanism (MHSAM), (2) Feature Regularization Mechanism (FRM), (3) Self-Attention Feature Fusion Module (SAFFM), and (4) Residual Learning Module (RLM). 
MHSAB outputs attention weights $\alpha(x)$, and \textbf{fusion features} $z(x)$ and $p^*(x)$ that capture key local information. 
These $\alpha(x)$, $z(x)$ and $p^*(x)$, will be refined in the Attention Competition Mechanism (ACM).

The ACM is composed of a series of loss functions and a regularization item; and it updates attention weights $\alpha(x)$, and \textbf{fusion features} $z(x)$ and $p^*(x)$ to enhance key person  information and suppress non-key  person  information. 

\textbf{In  the testing  stage.}  For the standard  person Re-ID task,  we concatenate the  feature  vector $p^{*}(x) \in \mathbb{R}^{512}$ and $q^{*}(x) \in \mathbb{R}^{512}$  to  find the best matching person in the gallery by comparing the squired distance,i.e. $d(a, b) =\|a-b\|^{2}_{2}$.  For the occlusion  person task,   we  only use  the feature  vector $p^{*}(x) \in \mathbb{R}^{512}$   to  find the best matching person in the gallery by comparing the squared distance.

\section{Global Feature Branch (Baseline)}

Following recent state-of-the-art methods~\cite{Kaiyang19, Yifan2018, Bryan2019, Wenjie19, Ruibing19, Zuozhuo19, Guanshuo2018}, 
we adopted ResNet-50 (pre-trained on ImageNet~\cite{Imagenet2009}) as the backbone network to encode a person image $x$. 
We modify the backbone ResNet-50 slightly to extract richer information via larger-sized high-level feature maps. 
The down-sampling operation at the beginning of stage $4$ is not employed, then the output of the Backbone Network is $Q(x) \in \mathbb{R}^{24 \times 8 \times 2048}$. Following~\cite{Zuozhuo19, Yifan2018, Guanshuo2018}, we also append a series of downsampling operations to the large feature map $Q(x)$. 
As shown in Fig.~\ref{Fig:Pipeline_MHSA}, firstly we employ a global average pooling operation on the output feature $Q(x)$, the $24 \times 8  \times 2048$ tensor from the stage $4$ of ResNet-50, to get a feature vector $q(x) \in \mathbb{R}^{2048}$. Then, $q(x)$ is further reduced to a $512$-dimensional feature vector $q(x)^{*}$  through a $1 \times 1$ convolution layer, a batch normalization layer, and a \textcolor{red}{Rectified Linear Units (ReLU)} layer. 
Finally, the feature vector $q(x)^{*}$ is fed into the loss function, which is cross entropy loss  $\mathcal{L}_{CE}$ in this baseline model.
The baseline  model  of  our  MHSA-Net  is   composed  of  the	Global Feature Branch  and the BackBone  Network.

Unlike existing methods~\cite{Kaiyang19, Bryan2019, Wenjie19, Ruibing19, ZHAO2020107014}, we don't introduce the triplet loss into the GFB. In our experiments, we observed that incorporating triplet loss in the GFB negatively impacts the performance of MHSA-Net on generic person Re-ID with occlusions. It seems this more strict constraint on global features affects the local feature capturing in some degree. So, in our MHSA-Net, the loss function in the baseline model only contains $\mathcal{L}_{CE}$.

\section{Multi-Head Self-Attention Branch (MHSAB)}

We introduce the Multi-Head Self-Attention Mechanism (MHSAM)~\cite{Ashish2017, 2020Jean-Baptiste, ZHAO2022360} into the person Re-ID pipeline, to help the network capture key local information from occluded images. 
However, there  are two issues   need to be solved  for this  MHSAM in  the  occluded person  Re-ID  task.

(1) MHSAM~\cite{Ashish2017,2020Jean-Baptiste} can capture key local information using multiple embeddings; but these existing methods directly concatenate these embeddings, which result in a huge dimensional feature space, making search and training expensive and difficult. For our Re-ID task which is more complicated than the Natural Language Processing (NLP) task in~\cite{Ashish2017, Zhouhan2017}, we need a more effective design on MHSAM  to output  a low  dimensional  and efficient person   descriptor for the  person matching task.

(2) MHSAM produces multiple attention maps and feature embeddings for an image to encode rich information, which enhances the robustness of the deep model in representation learning~\cite{Ashish2017}.  But this design itself often makes different embeddings to redundantly encode similar or same personal information. 
Thus, it is desirable to make the generated embeddings diverse, namely, they capture various features of the person from different aspects.

Based on these observations, we propose a novel Multi-Head Self-Attention Branch (MHSAB) to tackle the above issues. 
MHSAB contains three components: the Multi-Head Self-Attention mechanism (MHSAM), Feature Regularization Mechanism (FRM), and Self-Attention  Feature Fusion Module (SAFFM). 
The MHSAM computes multiple attention maps for key sub-regions and multiple embeddings for each person image.  
The FRM contains a Feature Diversity  Regularization Term (FDRT) and an Improved Hard Triplet Loss  (IHTL) function.
The FDRT enhances the diversity of the multiple embeddings in MHSAM, and the IHTL refines each individual embedding to better capture key information. 
The SAFFM adaptively combines multiple embeddings to produce a fused low-dimension feature vector. 

\subsection{Multi-Head Self-Attention Mechanism (MHSAM)}
As described in Section~\ref{Introduction}, it is desirable to adaptively capture key local features in unoccluded regions and avoid information from occluded regions. 
To achieve this, we adopt a Multi-Head Self-Attention mechanism (MHSAM)~\cite{Ashish2017,  2020Jean-Baptiste, DBLP01232}.  The  architecture  of  MHSAM  is  shown  in Fig.~\ref{Fig_MHSAM_1}.
Here,  we build $K$-head for this MHSAM in two steps: 
(1) first, given a person image, we learn its $K$ attention weights $\alpha(x) \in \mathbb{R}^{J \times K}$ (where $J=[24 \times 8]$) on each pixel $j \in J$ of feature maps $Q(x) \in \mathbb{R}^{J \times 2048}$. 
(2) Second, we compute the $K$ attention-weighted embeddings of $Q(x)$ for this person image. Specifically:

\textbf{(Step 1)} Compute $\alpha(x)$ by the Attention Weight Calculation module (Figs.~\ref{Fig_MHSAM_1} ):
\begin{equation}~\label{eq:alpha}
\alpha(x) = softmax(\omega_{2}ReLU(\omega_{1}Q(x)^{T})), 
\end{equation}
where $Q(x) \in \mathbb{R}^{J \times 2048}$ is reshaped to a matrix in $\mathbb{R}^{192 \times 2048}$, $\alpha(x) \in \mathbb{R}^{K \times 192}$ is reshaped to a tensor in $\mathbb{R}^{24 \times 8 \times K}$,  $\omega_{2} \in \mathbb{R}^{K \times 512}$ and $\omega_{1} \in \mathbb{R}^{512 \times 2048}$ are two parameter weight matrices to learn, and the softmax is applied pixel-wise so that on each pixel the $K$ attention weights sum up to one. 

\textbf{(Step 2)} Multiply the attention weight $\alpha(x)$ with feature maps $Q(x)$, and further apply a non-linear transformation, to get $K$ attention-weighted embeddings $P \in \mathbb{R}^{K \times 512}$ (Figs.~\ref{Fig_MHSAM_1}):  
\begin{equation}~\label{eq3}
P(x) = AvgPool(\alpha(x) \odot  Q(x))\omega_{3} + b_{3}, 
\end{equation}
where $\omega_{3} \in \mathbb{R}^{2048 \times 512}$ is the parameter weight matrix, $AvgPool(\cdot)$ is  the average  pooling operation,  and $b_{3} \in \mathbb{R}^{512}$ is the bias to learn for the fully connection layer ``FC'' in Fig.~\ref{Fig_MHSAM_1}.  Since this,  we can obtain  $K$  feature  branch  heads,  and  the number  of  the heads is $K$.  The $\odot$  is  the element-wise  product  operation.
 
The attention weights $\alpha(x)$ in Eq.~(\ref{eq:alpha}) 
are adaptively learned toward the objective of person matching in Re-ID. 
Greater $\alpha$ values indicate bigger importance of pixels/local regions and vice versa. 
As some examples shown in Figs.~\ref{Fig1} and \ref{Fig8}, key information from unoccluded regions can be captured by MHSAM, while occluded regions can be suppressed.   Here,  the hyper-parameter $K$  is  discussed in the Subsection~\ref{AbMHSAM}.

\subsection{Feature Regularization Mechanism (FRM)}~\label{FRT} FRM contains a Feature Diversity Regularization Term (FDRT) and an Improved Hard Triplet Loss (IHTL). The FDRT encourages the multiple embeddings $P(x)$ to cover more key local information from various respects. 
The IHTL refines the embeddings so that they individually can better serve person matching. 
FRM takes in $P(x) \in \mathbb{R}^{K \times 512}$, and outputs a new tensor $P^{\bot}(x) \in \mathbb{R}^{K \times 512}$. 
 
\subsubsection{Feature Diversity Regularization Term (FDRT)}~\label{FDRT} 
 
The $K$ embeddings directly produced by MHSAM tend to capture  similar/same person information redundantly. 
To avoid this, following~\cite{Ashish2017}, we also introduce the  Feature Diversity Regularization Term (FDRT) into MHSAM, to regularize the $K$ representations and enforce their diversity. 

The $K$ embeddings in MHSAM are not overcomplete ~\cite{Hongyu2019Diversity, Di2017Diversity}. So we can restrict the Gram matrix of $K$ embeddings to be close to an identity matrix under Frobenius norm. 
Firstly, we create a Gram matrix $G(x)$ of $P(x)$ by  $G(x)=P(x)P(x)^{T}$. Each element in $G(x)$ denotes the  correlation between $P(x)$. Here, $P(x)$ is normalized so that they are on an $L_2$ ball. 
Secondly, to enhance the diversity of the $K$ feature vectors in $P(x)$, we minimize the deviation of $G(x)$ from the identity matrix. 
Therefore, we define the Feature Diversity Regularization Term (FDRT) as 
\begin{equation}~\label{eq5}
\mathcal{L}_{FDRT} = \frac{1}{K^{2}}\left \| G(x) - I  \right \|_1, 
\end{equation}
where $G(x)$ is  the gram matrices of  $P(x)$,  and $I \in \mathbb{R}^{K \times K}$ is an identity matrix. 
With FDRT, the $K$ embeddings $P(x)$ are more diverse and can capture key information from different perspectives, which enhances the model robustness.

\subsubsection{Improved Hard Triplet Loss (IHTL)}
MHSAM produces $K$ embeddings $P(x) \in \mathbb{R}^{K \times H}$ for each person image.  
To further filter out non-key information, we design a new loss function to help train the network 
so that each individual embedding can be used separately for person matching. 
We are inspired by the  hard triple loss~\cite{Alexander2017}, which uses a hard sample mining strategy to achieve desirable performance. 
Hence, we propose an Improved Hard Triplet Loss (IHTL)  by  revising the  hard triple loss~\cite{Alexander2017}.

Before defining  the  Improved Hard Triplet Loss (IHTL),  we  firstly organize the training samples into a set of triplet feature units, $S = {(s(x^{a}), s(x^{p}), s(x^{n}))}$, or simply $S = {(s^{a}, s^{p}, s^{n})}$ in the following. 
The raw person image triplet units is $X = {(x^{a}, x^{p}, x^{n})}$. Here, $(s^{a}, s^{p})$ represents a positive pair of features $y^{a}= y^{p}$, and $(s^{a}, s^{n})$ indicates a negative pair of features with $y^{a} \neq y^{n}$. Here, $y\in Y$ is the person ID.

In  the  hard  triple  loss~\cite{Alexander2017}, a hard-sample mining strategy is  introduced: a positive sample pair with the largest distance is defined as the \emph{hard positive sample pair}; the negative sample pair with the smallest distance is defined as the \emph{hard negative sample pair}. 
The hard  triple  loss function can then be defined using hard sample pairs: 
\begin{equation}~\label{eq7}
\mathcal{T}_{HardTriplet}= ln(1+ exp(\underset{x^{a},x^{p}}{\text{max}} d(s^{a}, s^{p})- \underset{x^{a},x^{n}}{\text{min}} d(s^{a}, s^{n}))),  
\end{equation}

Based  on the hard  triplet  loss  function,   we define an Improved Hard Triplet  Loss (IHTL). 
We define the \emph{improved hard positive sample pair} and \emph{improved hard negative sample pair} in two steps:  \textbf{(I):} 
Between each sample image pair, $K \times K$ distances can be computed, because each person image has $K$ embeddings $P(x) \in \mathbb{R}^{K \times 512}$ in MHSAM. 
We use the largest distance from these distances to measure the embeddings of the positive sample pairs, and use the smallest distance from these distances for the negative sample pairs.

\textbf{(II):} We further use the hard samples mining strategy ~\cite{Alexander2017} to define the hard sample pairs. The improved  hard  positive sample  pair   is  $\underset{x^{a},x^{p}}{\text{max}} \underset{i,j}{\text{max}} \, d(P(x^{a})_{i}, P(x^{p})_{j})$;  the improved  hard  negative sample  pair  is   $\underset{x^{a},x^{n}}{\text{min}} \underset{i,j}{\text{min}} \, d(P(x^{a})_{i}, P(x^{n})_{j})$.   The  Improved Hard Triplet loss is defined as:
\begin{equation}~\label{eq8}
\begin{aligned}
\mathcal{T}_{IHTL} = ln(1+ exp(& \underset{x^{a},x^{p}}{\text{max}} \underset{i,j}{\text{max}} \; d(P(x^{a})_{i}, P(x^{p})_{j})-\\
&\underset{x^{a},x^{n}}{\text{min}} \underset{i,j}{\text{min}} \; d(P(x^{a})_{i}, P(x^{n})_{j}))),
\end{aligned}
\end{equation}
where $i,j \in \{1,2, \cdots, K \}$, $d(a, b) =\|a-b\|^{2}_{2}$ denotes the squired distance in feature space.  Here,   
During training, the IHTL  refines embeddings so that they individually can perform better person matching. This encourages the embeddings to focus on important information.

\subsection{Self-Attention Feature Fusion Module (SAFFM)}~\label{SFFM}

The output of \textbf{FDRT}, the $K$ embeddings $P^{\bot}(x) \in \mathbb{R}^{K \times 512}$ covers various properties of a person image. But directly using $P^{\bot}(x)$ by concatenation will lead to dimension explosion in person matching. 
Thus, we design a Self-Attention Feature Fusion Module (SAFFM) to first learn $K$  attentional weights by a series  of neural networks, then fuse $P^{\bot}(x)$ to get a lower-dimensional $p(x)^{*} \in \mathbb{R}^{512}$.

\begin{figure}\centering
	\begin{center} 
		\includegraphics[scale=0.26]{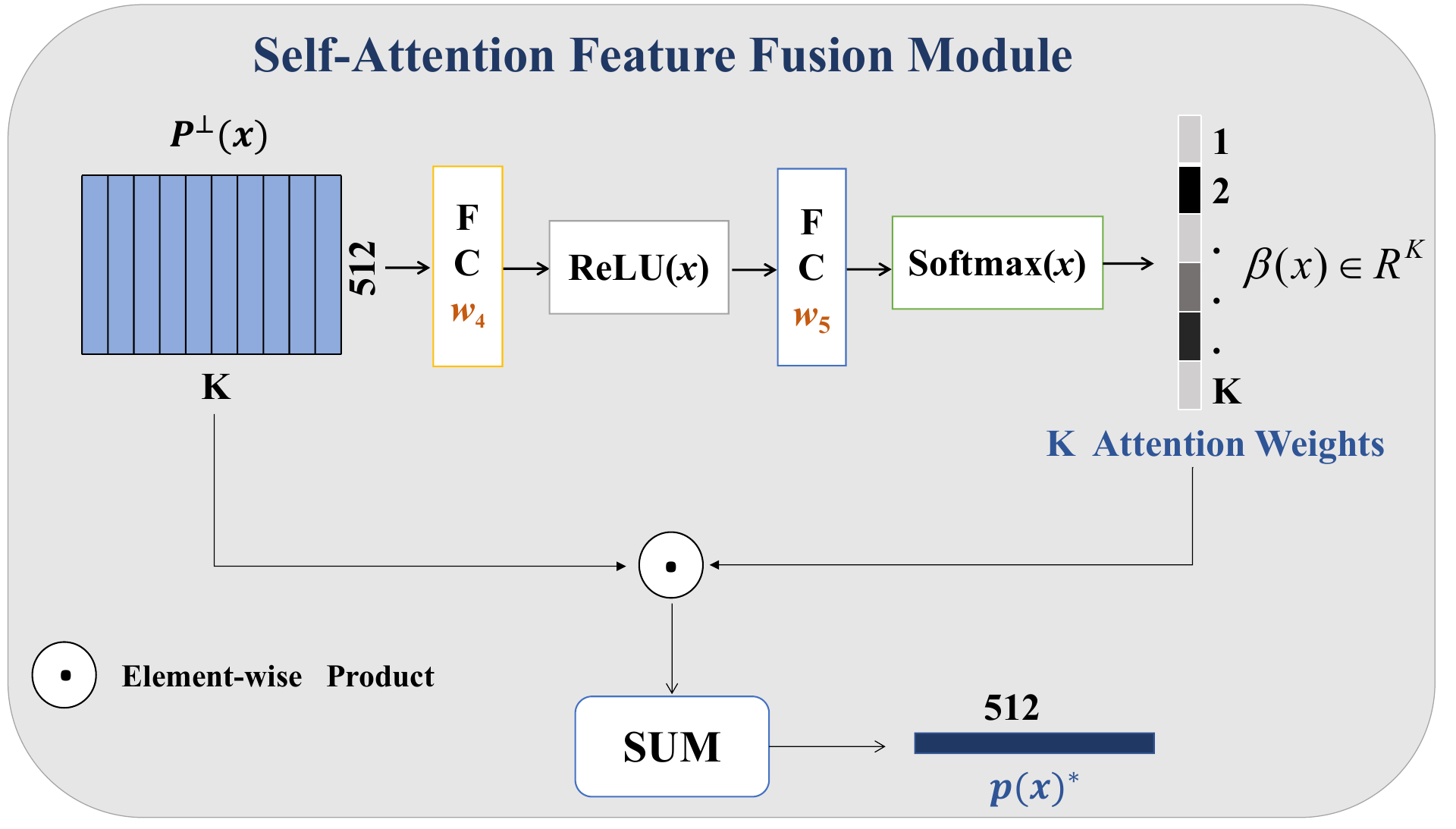}
	\end{center} 
	\caption{The Self-Attention Feature Fusion Module.} 
	\label{Fig:SAFFM} 
\end{figure}

Specifically, \underline{Step-1}, compute the attentional  weight $\beta(x) \in \mathbb{R}^{K \times  512}$ (Fig.~\ref{Fig:Pipeline_MHSA}  and  Fig.~\ref{Fig:SAFFM}). 
The  matrix $P^{\bot}(x) \in \mathbb{R}^{K \times 512}$  is transposed to $P^{*}(x) \in \mathbb{R}^{512 \times K}$, then compute $\beta$ by
\begin{equation}~\label{eq10}
\beta(x) = softmax(\omega_{5}ReLU(\omega_{4}P^{*}(x))), 
\end{equation}
where $\omega_{4} \in \mathbb{R}^{512 \times 1024}$ and  $\omega_{5} \in \mathbb{R}^{1024 \times 512}$  are  two parameter weight matrices to learn, and the softmax is applied pixel-wise so that each pixel on the  each  attention  vector  of  the $\beta(x)$ sum up to one.  

\underline{Step-2}, compute the self-attention weighted feature  vector $p(x)^{*} \in \mathbb{R}^{512}$, by 
\begin{equation}~\label{eq11}
p(x)^{*}= \sum_{i=1}^K [\beta(x) \odot P^{\bot}]_i. 
\end{equation}
Here, $p(x)^{*} \in \mathbb{R}^{512}$,  $\odot$  is  the element-wise  product  operation.

SAFFM reduces the dimension of the multiple  embeddings $P^{\bot}(x)$ for both training and testing. 
In the training stage,  $p^{*}(x)$ is also fed to the cross entropy loss and the hard triplet loss function, 
\begin{equation}~\label{eq12}
\mathcal{L}_{SAFFM} =\mathcal{L}_{CE}^{*}+\mathcal{T}_{HardTriplet}^{*}.
\end{equation}
Here, the input to both $\mathcal{L}_{CE}^{*}$ and $\mathcal{L}_{HardTriplet}^{*}$ is $p^{*}(x)$.

\subsection{Residual Learning Module}
As shown in the RLM module in Fig.~\ref{Fig:Pipeline_MHSA}, 
with MHSAM, $P^{\bot}(x)$ aims to capture key  information in unoccluded regions from local  perspective; 
while $q(x)^{*}$ captures global information of the whole person image. 
To prevent $P^{\bot}(x)$ from being redundant with $q(x)^{*}$, we cast their feature fusion as a residual learning task. 
Specifically, 
(1) to match with the dimension $K \times 512$ of $P^{\bot}(x)$, we copy $q(x)^{*}$ for $K$ times to obtain $Q(x)^{*} \in \mathbb{R}^{K \times 512}$.
(2) 
The input to the residual block includes global feature $Q(x)^{*}$ and local feature $P^{\bot}(x)$. The parameters ($\omega_{1}, \omega_{2}, \omega_{3}, b_{3}$) of $P^{\bot}(x)$ will be optimized. 
(3) We define the residual  learning  embedding  as   
\begin{equation}~\label{eq4}
Z(x) = Norm(Q(x)^{*} + P^{\bot}(x)),
\end{equation}
where $Norm(\cdot)$ denotes the layer normalization~\cite{Jimmynormalization2016}. 
This RLM encourages $P^{\bot}(x)$ to only capture important local information. 

In the training stage, $Z(x) \in \mathbb{R}^{K \times 512}$ is simply summed along the first dimension to obtain $z(x) \in \mathbb{R}^{512}$. 
And $z(x)$ is also fed into the cross entropy loss and the  hard triplet loss function, i.e. 
\begin{equation}~\label{eq19}
\mathcal{L}_{ReN} =\mathcal{L}_{CE}^{**}+\mathcal{L}_{HardTriplet}^{**}
\end{equation}
Here, the input of $\mathcal{L}_{CE}^{**}$ and $\mathcal{L}_{HardTriplet}^{**}$ is $z(x)$. And $z(x)$ does not participate in person matching in the testing stage.

Finally, the \textbf{loss functions in MHSAB} are summarized as 
\begin{equation}~\label{eq13}
\mathcal{L}_{MHSAB} = \mathcal{L}_{SAFFM}+\lambda_1 \mathcal{L}_{FDRT} + \mathcal{L}_{ReN} + \lambda_2 \mathcal{T}_{IHTL},
\end{equation}
where $\lambda_1$ and $\lambda_2$ are the balance  paremeters (see detail in Subsection~\ref{AbFDRT} and the Subsection~\ref{AbIHT}).

\section{Attention Competition Mechanism}
MHSAB enhances attention on key sub-regions, but the extracted attention maps still contain some non-key information. We propose an Attention Competition Mechanism (ACM) to further refine the attention weights.

In \cite{thc2019}, an attention competition strategy was proposed to filter out attention information  of the  non-key words  in the  Text-to-Image generation task. 
This idea was composed  of  an attention regularization term  and a  series of  cross-modal  matching loss  functions. 
This has been shown effective in the Text-to-Image generation task. 
In  the image generation: an attention regularization term  can effectively  filter out  the attention information of non-key words; the    cross-modal  matching loss  functions can effectively enhance or preserve the attention information of the key words according to the objective.  
Similarly, we believe it  can also help  the person Re-ID model  filter  out the attention information of  the non-key sub-regions from the person images.  
Therefore, we also design a similar strategy in this Person Re-ID pipeline. 
To our knowledge, this is the first time a competition strategy was designed for Re-ID task.
Through  a  series  of experiments, we observe that this mechanism is promising.

\begin{figure}
   \centering
   \subfigure[Initial Stage.]{\includegraphics[scale=0.3]{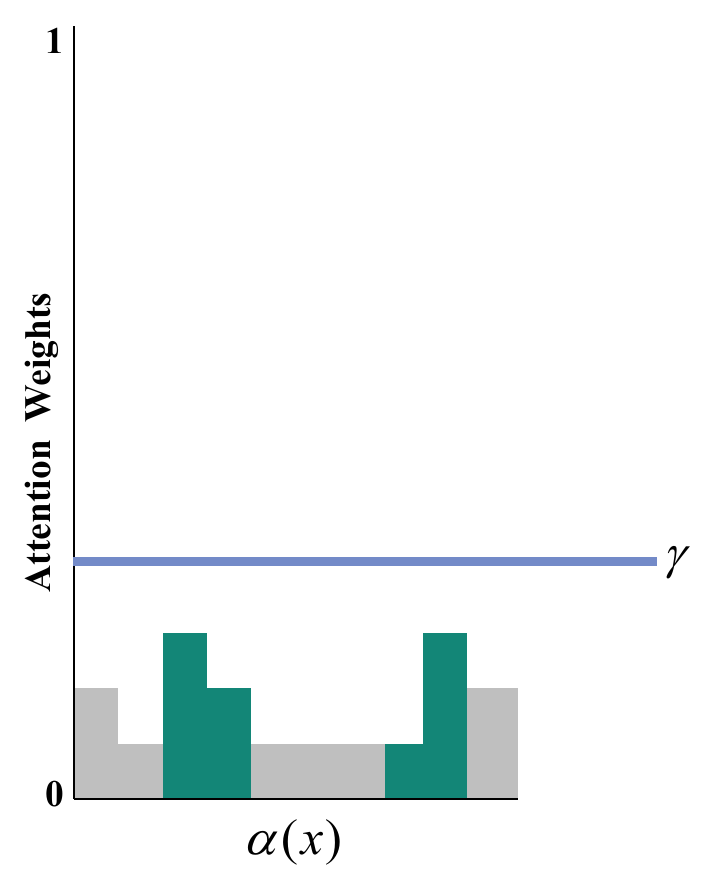}}
   \subfigure[After ACM Refinement.]{\includegraphics[scale=0.3]{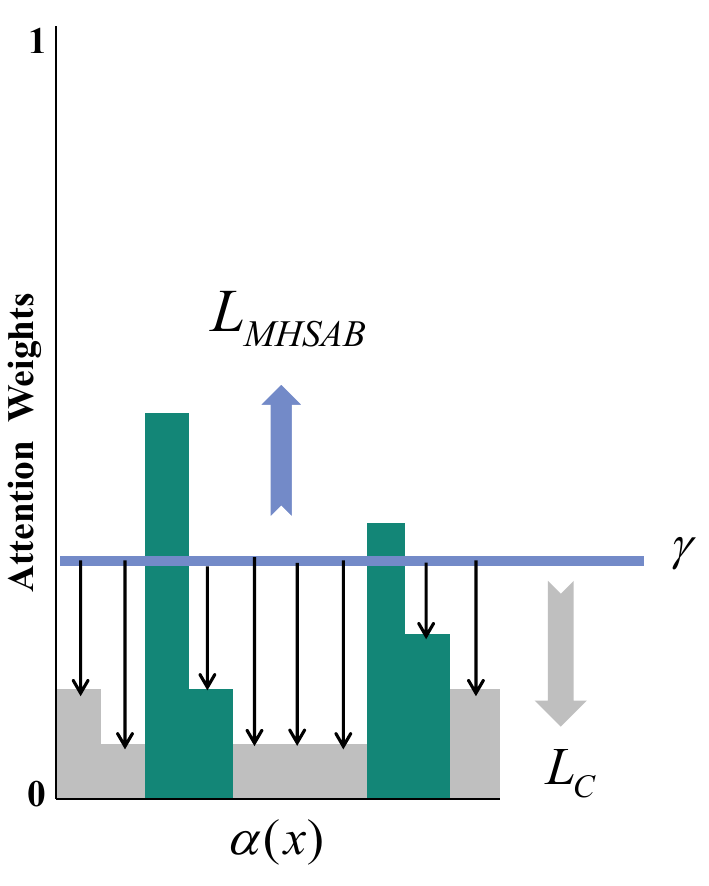}}
   \caption{The schematic diagram of the attention  competition process on $\alpha(x)$. Grey columns are the attention weights of non-key sub-regions, and green columns are the weights of key sub-regions.}
   \label{Fig:ACM} 
\end{figure}

Specifically, 
we use an attention regularization term to suppress non-key information, and use the aforementioned person Re-ID loss function $\mathcal{L}_{MHSAB}$ to enhance attention on important regions. 
The attention regularization term~\cite{thc2019} is defined as: 
\begin{equation}\label{eq14}
\mathcal{L}_{C}= \sum_{i,j} (\min(\alpha_{i,j}, \gamma))^{2},
\end{equation}
where the subscript ``C'' stands for ``\emph{Competition}'', and $\gamma > 0 $ is a threshold.
Fig.~\ref{Fig:ACM} shows a schematic diagram of the ACM. The grey columns illustrate attention weights on non-key sub-regions, and the green columns are for weights on key regions. 
In the initial state of training, as shown in sub-fig (a), all attention weights in $\alpha$ are small. 
In ACM, the attention regularization term $\mathcal{L}_{C}$ sets a threshold and pushes the attention weights lower than this threshold toward zero; while $\mathcal{L}_{MHSAB}$ increases attention weights of sub-regions if they benefit person matching. An  illustration of this procedure is given in (b).

\medskip

\textbf{The total loss functions in MHSA-Net.} 
The total loss function $\mathcal{L}_{Total}$  is 
\begin{equation}~\label{eq15}
\mathcal{L}_{Total}=\mathcal{L}_{MHSAB}+\mathcal{L}_{GFB}+\lambda_3\mathcal{L}_{C},
\end{equation}
where $\lambda_3$ is the balance parameter (see its discussion in the Section~\ref{ACM_ablation}).

\section{Experiment}
To evaluate the MHSA-Net, we conduct extensive experiments on three widely used generic person Re-ID benchmarks, i.e. \textbf{Market-1501}~\cite{Zheng2015Scalable}, \textbf{DukeMTMC-reID}~\cite{Ristani2016Performance, ZhedongZheng} and \textbf{CUHK03}~\cite{Weireid2014} datasets, 
and four occluded person Re-ID benchmarks, i.e. \textbf{Occluded-DukeMTMC}~\cite{Jiaxu112019},  \textbf{P-DukeMTMC-reID}~\cite{Jiaxuan2018}, \textbf{Partial-REID}~\cite{WeiShi2015}   and  \textbf{Partial-iLIDS}~\cite{Lingxiao2018}. 
First, we compare the performance of MHSA-Net with state-of-the-art methods on these datasets. 
Second, we perform ablation studies to validate the effectiveness of each component. 

\subsection{Datasets and Evaluation}
We  follow  almost all person Re-ID approaches~\cite{Zhun2020, Kaiyang19, Zuozhuo19, Yifan2018, Ristani2016Performance, ZhedongZheng, Lingxiao2018, WeiShi2015, Jiaxu112019, Jiaxuan2018, 8099841, ZHAO2019161}  to set  the following seven  person Re-ID  datasets.

\textbf{Market-1501}~\cite{Zheng2015Scalable} has $32,668$ labeled images of $1,501$ identities collected from $6$ camera views. The dataset is partitioned into two non-overlapping parts: the training set with $12,936$ images from $751$ identities, and the test set with $19,732$ images from $750$ identities. 
In the testing stage,  $3,368$ query images from $750$ identities are  used to retrieve the persons from the rest of the test set, i.e. the gallery set. 

\textbf{DukeMTMC-reID}~\cite{Ristani2016Performance, ZhedongZheng} is another large-scale person Re-ID dataset. It has $36,411$ labeled images of $1,404$ identities collected from $8$ camera views. The training set consists of $16,522$ images from $702$ identities; We used $2,228$ query images from the other $702$ identities, and $17,661$ gallery images.

\textbf{CUHK03}~\cite{Weireid2014} is a challenging Re-ID benchmark. It has $14,096$ images of $1,4674$ identities captured from $6$ cameras. It contains two datasets. \textbf{CUHK03-Labeled:} the bounding boxes of person images are from manual labeling. 
\textbf{CUHK03-Detected:} the bounding boxes of person images are detected from deformable part models (DPMs), which is more challenging due to severe bounding box misalignment and background cluttering.  
Following \cite{Zhun2020, Kaiyang19, Zuozhuo19, Yifan2018}, we used the $767/700$ split~\cite{Weireid2014} of the detected images.

\textbf{Occluded-DukeMTMC}~\cite{Jiaxu112019} has $15,618$ training images, $17,661$ gallery images, and $2,210$ occluded query images. We use this dataset to evaluate our  MHSA-Net in Occluded Person Re-ID task.

\textbf{P-DukeMTMC-reID}~\cite{Jiaxuan2018} is a modified version based on DukeMTMC-reID~\cite{Ristani2016Performance, ZhedongZheng}. There are $2,652$ images ($665$ identifies) in the  training set, $2,163$ images ($634$ identities) in the  query set and $9,053$ images in the gallery set.

\textbf{Partial-REID}~\cite{WeiShi2015} is a specially designed partial person Re-ID benchmark that has $600$ images from $60$ people. Each person has five partial images in query set and five full-body images in gallery set. 
These images are collected at a university campus under different viewpoints, backgrounds, and occlusions.

\textbf{Partial-iLIDS}~\cite{Lingxiao2018} is a simulated partial person Re-ID dataset based on the iLIDS dataset. It  has a total of $476$ images of $119$ people.

\textbf{Evaluation Protocol.} We employed two standard metrics adopted in most person Re-ID approaches, namely, the cumulative matching curve (CMC) that generates ranking accuracy, and the mean Average Precision (mAP).  The CMC curve shows the probability that a query identity appears in different-sized candidate lists. This evaluation measurement is valid only if there is only one ground truth match for a given query. In this paper,  we  report  the  Rank-1 accuracy. 
The mAP  calculates the area under the Precision-Recall curve, which is known as average precision (AP). Then, the mean value of APs of all queries, i.e.,
mAP, is calculated, which considers both precision and recall of an algorithm, thus providing a more comprehensive evaluation.

\begin{table*}[h!tbp]
	\centering
	\caption{The comparison with the many state-of-the-art generic person re-ID methods on Market-1501, DukeMTMC-reID, and CUHK03 datasets.      	MHSA-Net$^{\dagger}$ indicates that we  drop  the $\mathcal{L}_{CE}$ in the training  process.
		MHSA-Net$^{*}$  indicates that we only use the key local feature $p^{*}(x)$ from MHSAB+ACM to conduct the person matching  task. }
	
	\setlength{\tabcolsep}{4mm}{
		\begin{tabular}{c| c c | c c |  c  c | c  c }
			\hline \hline
			\multirow{2}{*}{Method}   
			&\multicolumn{2}{|c|}{Market-1501} &\multicolumn{2}{|c}{DukeMTMC-reID} &\multicolumn{2}{|c}{CUHK03-Detected} &\multicolumn{2}{|c}{CUHK03-Labeled}  \\
			\cline{2-9} 	
			& Rank-1 & mAP & Rank-1 & mAP   & Rank-1 & mAP   & Rank-1 & mAP \\	
			\hline

			MLFN \cite{Xiaobin18}   & $90.0$ & $74.3$ & 81.0  & 62.8  & 52.8  & 47.8   & 54.7   & 49.2   \\ 
			HA-CNN~\cite{li2018harmonious}   &  91.2   & 75.7   & 80.5   & 63.8   & 41.7  & 38.6   &  44.4  & 41.0  \\
			PCB+RPP\cite{Yifan2018}   & 93.8    & 81.6     &    83.3    &   69.2     & 62.8    & 56.7  & - & - \\
			Mancs\cite{cheng2018eccv}     &   93.1    & 82.3    &   84.9    & 71.8    &   65.5    & 60.5  \\ 
			PAN~\cite{Zheng_2018}   & 82.8  & 63.4    & 71.6  & 51.5   & 36.3  & 34.0  & 36.9  & 35.0  \\ 
			FANN\cite{Discriminative_2019}   & 90.3    & 76.1   & -& - & 69.3    & 67.2    & -& - \\
			VCFL\cite{Fangyi19}     &   90.9  & 86.7       & - & -   &   70.4  & 70.4  &  - &  - \\
			PGFA~\cite{Jiaxu112019}   &   91.2    & 76.8   &   82.6    & 65.5    &   -   & -  & -  & - \\    
			HACNN+DHA-NET~\cite{ZhengWang2020}  & 91.3    & 76.0   & 81.3   & 64.1   & - & - & - & - \\
			IANet\cite{Ruibing19}     &   94.4    & 83.1  &   87.1    & 73.4   & - & -  & - & - \\ 
			BDB\cite{Zuozhuo19}     &   94.2    & 84.3   &   86.8    & 72.1   &  72.8    & 69.3  & 73.6  & 71.7 \\
			AANet\cite{AANET19}     &   93.9    & 83.4    &   87.7    & 74.3   & - & - & - & - \\ 
			CAMA\cite{Wenjie19}      &   94.7   & 84.5   &   85.8    & 72.9    &   66.6    & 64.2  \\ 
			OSNet\cite{Kaiyang19}   &   94.8    & 84.9   &   88.6    & 73.5    &   72.3    & 67.8   & -  & - \\  
			RANGEv2\cite{WU2022108239}   &   94.7    & 86.8   &  87.0    & 78.2    &   64.6    & 61.6   & 67.4    & 64.3 \\  
			JWSAA\cite{NING2021801}  &   94.8    & 83.2   &   88.3    & 75.6    &   72.3    & 67.8   & -  & - \\  	
			HOReID~\cite{Guanan2020}   &   94.2   & 84.9   &  86.9    & 75.6    &   -  & -  & -  & - \\  
		    SCSN (4-stage)~\cite{9156982}   &   92.4   & 88.3   &  91.0    & 79.0    &   84.7  & 81.0 & 86.8  & 84.0 \\  
		    SCSN (3-stage)~\cite{9156982}   &   95.7  & 88.5   &  90.1    & 79.0    &   84.1  & 80.2 & 86.3 & 83.3 \\ 
		    RGA-SC~\cite{9157488}   &   95.8  & 88.1   &  86.1    & 74.9    &   77.3  & 73.3 & 80.4 & 76.4 \\ \hline
			Baseline      &   92.0    & 78.8    &   81.0    & 62.8   &   56.3    & 53.0   &   58.6    & 55.2   \\
			MHSA-Net (K=5)         & 94.3  & 83.5          &    87.1  &  73.0         &    73.4  & 70.2         &  75.8  &  73.0  \\
			MHSA-Net (K=8)         & 94.6   & 84.0         &   87.3   & 73.1        &     72.8  & 69.3      &  75.6  &  72.7   \\ 
			MHSA-Net$^{*}$ (K=8)  & 94.0  & 82.9  &  86.3   &  72.5      &    72.4   & 69.7  &  74.4  & 72.0   \\   
			MHSA-Net$^{\dagger}$ (K=8)  & 94.3  & 82.5  &   87.0   &  72.6      &    72.7   & 69.9  &  75.2  & 72.3   \\  
			MHSA-Net+Re-ranking~\cite{Zhong2017Re} (K=8)           &  95.5  &  93.0       &  90.7  & 87.2        &    80.2  & 80.9     &  82.6  &  84.2  \\    \hline
			
			\hline \hline
	\end{tabular} }
	\label{SOTA-1} 
\end{table*}

\begin{table}[h!t]
	\centering
	\caption{ The comparison with the other occluded  person re-ID methods on Occluded-DukeMTMC dataset. 	
		MHSA-Net$^{\dagger}$ indicates that we  drop  the $\mathcal{L}_{CE}$ in the training  process. 
		MHSA-Net$^{*}$  indicates that we only use the key local feature $p^{*}(x)$ from MHSAB+ACM to conduct the person matching  task. 
		The \textcolor{red}{first}, \textcolor{ForestGreen}{second} and \textcolor{blue}{third} highest scores are shown in \textcolor{red}{red}, \textcolor{ForestGreen}{green} and \textcolor{blue}{blue} respectively.}
	
	\setlength{\tabcolsep}{2.5mm}{
		\begin{tabular}{c| c c c  c}
			\hline \hline
			\multirow{2}{*}{Method}   &
			\multicolumn{4}{|c}{Occluded-DukeMTMC}  \\
			\cline{2-5} 	
			&   Rank-1 & Rank-5 &  Rank-10 &  mAP  \\	
			\hline
			Random Erasing~\cite{2017Zhong} & 40.5  & 59.6  &  66.8  & 30.0 \\
			HA-CNN~\cite{li2018harmonious} & 34.4  & 51.9   &  59.4   & 26.0  \\
			Adver Occluded~\cite{2018Adversarially} &  44.5  &  - & - & 32.2  \\
			PCB~\cite{Yifan2018} &  42.6    &  57.1   &  62.9    & 33.7   \\  
			Part Bilinear~\cite{201xf8Yumin} &  36.9    & - &  - &  -  \\
			FD-GAN~\cite{Yixiao2018} &  40.8    & - &  - &  -  \\  
			DSR~\cite{Lingxiao2018} & 40.8   &  58.2   &  65.2   & 30.4   \\
			SFR~\cite{Lingxiao2018xl} &  42.3   &  60.3    &  67.3   &  32.0   \\
			PGFA~\cite{Jiaxu112019} &  51.4  &  68.6   & 74.9   &  37.3   \\
			HOReID~\cite{Guanan2020}  &  55.1   &  -  & -  &  \textcolor{blue}{43.8 } \\ 
			PVPM+Aug~\cite{Pose-guided2020}  &   57.3  &  72.6  &   77.2   &  \textcolor{red}{45.7 }   \\   \hline
			Baseline  & 38.9  &  53.5  & 60.1   & 25.6  \\
			MHSA-Net$^{*}$  (K=8)  &  \textcolor{red}{59.7 } &  \textcolor{red}{74.3 } & \textcolor{red}{79.5 } & \textcolor{ForestGreen}{44.8 }  \\
			MHSA-Net (K=8)  &  \textcolor{blue}{55.4 } &  \textcolor{blue}{70.2 } & \textcolor{blue}{76.4 }  & 42.4  \\
			MHSA-Net$^{\dagger}$ (K=8)  &  \textcolor{ForestGreen}{58.2 } &  \textcolor{ForestGreen}{73.2 } & \textcolor{ForestGreen}{78.4 }  & 43.1  \\ 
			\hline \hline
	\end{tabular} }
	\label{SOTA-2} 
\end{table}

\begin{table}[h!t]
	\centering
	\caption{ The comparison with the other occluded  person re-ID methods on P-DukeMTMC-reID dataset. 	
		MHSA-Net$^{\dagger}$ indicates that we  drop  the $\mathcal{L}_{CE}$ in the training  process.
		MHSA-Net$^{*}$  indicates that we only use the key local feature $p^{*}(x)$ from MHSAB+ACM to conduct the person matching  task.
		The \textcolor{red}{first}, \textcolor{ForestGreen}{second} and \textcolor{blue}{third} highest scores are shown in \textcolor{red}{red}, \textcolor{ForestGreen}{green} and \textcolor{blue}{blue} respectively.}
	
	\setlength{\tabcolsep}{2.5mm}{
		\begin{tabular}{c| c c c  c}
			\hline \hline
			\multirow{2}{*}{Method}   &
			\multicolumn{4}{|c}{P-DukeMTMC-reID}  \\
			\cline{2-5} 	
			&   Rank-1 & Rank-5 &  Rank-10 &  mAP  \\	
			\hline
			OSNet\cite{Kaiyang19} & 33.7   &  46.5   &   54.0   &  20.1   \\
			PCB+RPP\cite{Yifan2018} &  40.4   &  54.6   &  61.1   &  23.4    \\
			PCB\cite{Yifan2018}  &  43.6   &   57.1   &  63.3  &   24.7  \\
			PGFA~\cite{Jiaxu112019} &    44.2   &   56.7   &   63.0   &   23.1  \\ 
			PVPM+Aug~\cite{Pose-guided2020}  &   51.5  &  64.4  &   69.6  &  29.2   \\   \hline
			Baseline  & 61.0  &  72.5  & 78.4   & 27.0  \\
			MHSA-Net$^{*}$  (K=8)  &  \textcolor{red}{70.7 } &  \textcolor{ForestGreen}{81.0 } & \textcolor{ForestGreen}{84.6 } & \textcolor{red}{41.1 } \\
			MHSA-Net (K=8)  &  \textcolor{blue}{67.9 } &  \textcolor{blue}{79.7 } & \textcolor{blue}{83.7 }  & \textcolor{ForestGreen}{37.6 } \\
			MHSA-Net$^{\dagger}$ (K=8)  &  \textcolor{ForestGreen}{69.6 } &  \textcolor{red}{81.4 } & \textcolor{red}{85.0 }  & \textcolor{blue}{37.5 } \\ 
			\hline \hline
	\end{tabular} }
	\label{SOTA-4} 
\end{table}

\begin{table}[h!t]
	\centering
	\caption{The comparison with the other occluded  person re-ID methods on Partial-REID and PartialiLIDS. 
		MHSA-Net$^{\dagger}$ indicates that we  drop  the $\mathcal{L}_{CE}$ in the training  process.
		MHSA-Net$^{*}$  indicates that we only use the key local feature $p^{*}(x)$ from MHSAB+ACM to conduct the person matching  task.
		The \textcolor{red}{first}, \textcolor{ForestGreen}{second} and \textcolor{blue}{third} highest scores are shown in \textcolor{red}{red}, \textcolor{ForestGreen}{green} and \textcolor{blue}{blue} respectively. }
	\setlength{\tabcolsep}{3mm}{
		\begin{tabular}{c|c c | c c}
			\hline \hline
			\multirow{2}{*}{Method}   &
			
			\multicolumn{2}{|c|}{Partial-REID} &\multicolumn{2}{|c}{Partial iLIDS}  \\
			\cline{2-5} 	
			&  Rank-1 & Rank-3 & Rank-1 & Rank-3  \\	
			\hline
			MTRC~\cite{Shengcai2013}    &   23.7    &  27.3      &  17.7   & 26.1  \\ 	
			AMC+SWM~\cite{WeiShi2015}  &   37.3    &  46.0      &  21.0   & 32.8  \\
			DSR~\cite{Lingxiao2018}   &   50.7    &  70.0      &  58.8   & 67.2  \\
			SFR~\cite{Lingxiao2018xl}    &   56.9     & 78.5   &  63.9   & 74.8  \\
			FPR~\cite{LingxiaoHE19}  &  81.0     &  -    &  68.1  & - \\  
			PGFA~\cite{Jiaxu112019}    &    68.0     &   80.0    &   69.1  &  80.9  \\ 
			PVPM+Aug~\cite{Pose-guided2020}  &   80.6  &  84.2  &   68.7  & 81.4   \\  	 
			HOReID~\cite{Guanan2020}    &   \textcolor{blue}{85.3 }    &  \textcolor{ForestGreen}{91.0 }   &  72.6  & 86.4  \\   	\hline     
			Baseline  &   68.8     &  81.7    &  66.4  & 79.0  \\ 
			MHSA-Net$^{*}$ (K=8)   &   \textcolor{red}{85.7 }    &  \textcolor{red}{91.3 }   &  \textcolor{red}{74.9 } & \textcolor{red}{87.2 } \\ 
			MHSA-Net$^{\dagger}$ (K=8)   &   \textcolor{ForestGreen}{85.5 }   &  \textcolor{ForestGreen}{91.0 }   &  \textcolor{ForestGreen}{74.1 } & \textcolor{ForestGreen}{86.6 } \\
			MHSA-Net (K=8)   &   81.3     &  \textcolor{blue}{87.7 }   &  \textcolor{blue}{73.6 }  &  \textcolor{blue}{85.4 } \\             
			\hline\hline
	\end{tabular} }
	\label{SOTA-3} 
\end{table}

\subsection{Implementation Details}

Following many recent approaches~\cite{Jiaxu112019, Kaiyang19, Yifan2018, Zuozhuo19, AANET19}, the input images are re-sized to $384 \times 128$ and then augmented by random horizontal flip and  normalization  in the training stage. 
In the testing stage, the images are also re-sized to $384 \times 128$ and augmented only by normalization. Using the ImageNet pre-trained ResNet-50 as the backbone, our network is end-to-end in the whole training stage. 
Our network is trained using  $2$ GTX $2080$Ti  GPUs with a batch size of $128$. Each batch contains $32$ identities, with $4$ samples per identity. 
We use  Adam optimizer~\cite{2015Diederik} with $400$ epochs.  The base learning rate is initialized to $10^{-3}$ with a linear warm-up~\cite{PriyaGoyal2017} in first $50$ epochs, then decayed to $10^{-4}$ after $200$ epochs, and further decayed to $10^{-5}$ after $300$ epochs.  The whole training procedure has $400$ epochs and takes approximately $2$ hours.   Our MHSA-Net achieves  the  satisfactory performance in the  general  person Re-ID  and occluded  person Re-ID  tasks,  when $\lambda_1=1e-4$,  $\lambda_2=1.0$,  $\lambda_3=1e-3$,   $\gamma=1e-3$ and $K=8$.

\subsection{Comparison with state-of-the-art Methods}

In this subsection, we  compared MHSA-Net  with   a  series of state-of-the-art approaches  on  seven  person Re-ID  datasets.  
Here, MHSA-Net  concatenates the local  feature $p(x)^{*}$   and  global  feature $q(x)^{*}$   to conduct  the person  matching task. 
Compared with the proposed  MHSA-Net,  MHSA-Net$^{\dagger}$ indicates that we  drop  the $\mathcal{L}_{CE}$ in the training  process.  
The  MHSA-Net$^{*}$ only uses the local  feature $p(x)^{*}$  to conduct  the person Re-ID task.

\textbf{Person Re-ID on General Datasets.}
Firstly, we compared MHSA-Net with the state-of-the-art  generic  person Re-ID  approaches on Market-1501, DukeMTMC-Re-ID,  CUHK03-Labeled, and CUHK03-Detected  datasets,  and reported the results in  Tables~\ref{SOTA-1}. 
We randomly set $K=5$ and $K=8$ in these experiments (Through experiments, we observed that the $K$ value does not affect the result much. Some discussions on different $K$ values are given in Subsection~\ref{AbMHSAM}).  
Our MHSA-Net gets \emph{Rank-1= 94.6 ,  87.3 ,   73.4 ,  75.8    and mAP= 84.0 ,  73.1 , 70.2 ,  73.0   for Market-1501, DukeMTMC-reID, CUHK03-Detected  and CUHK03-Labeled}, respectively. 
If we introduce the Re-ranking~\cite{Zhong2017Re} into the MHSA-Net, i.e. MHSA-Net+Re-ranking (K=8), the accuracy further increases to \emph{Rank-1= 95.5 ,  90.7 ,   80.2 ,  82.6    and mAP= 93.0 ,  87.2 , 80.9 ,  84.2   for Market-1501, DukeMTMC-reID, CUHK03-Detected  and CUHK03-Labeled}, respectively. 
Recently, the state-of-the-art performance on Market-1501 and DukeMTMC-Re-ID has been saturated. 
Yet the MHSA-Net still gains effective improvement over the baseline model and outperforms most existing methods.  
The CUHK03 is the most challenging dataset among the three. Following the data setting in \cite{Zhun2020, Kaiyang19, Zuozhuo19, Yifan2018}, MHSA-Net also  outperforms the  most  state-of-the-art   methods on both CUHK03-Labeled and CUHK03-Detected datasets. 

\begin{figure}[h]
	\centering
	\begin{center} 
		\includegraphics[scale=0.3]{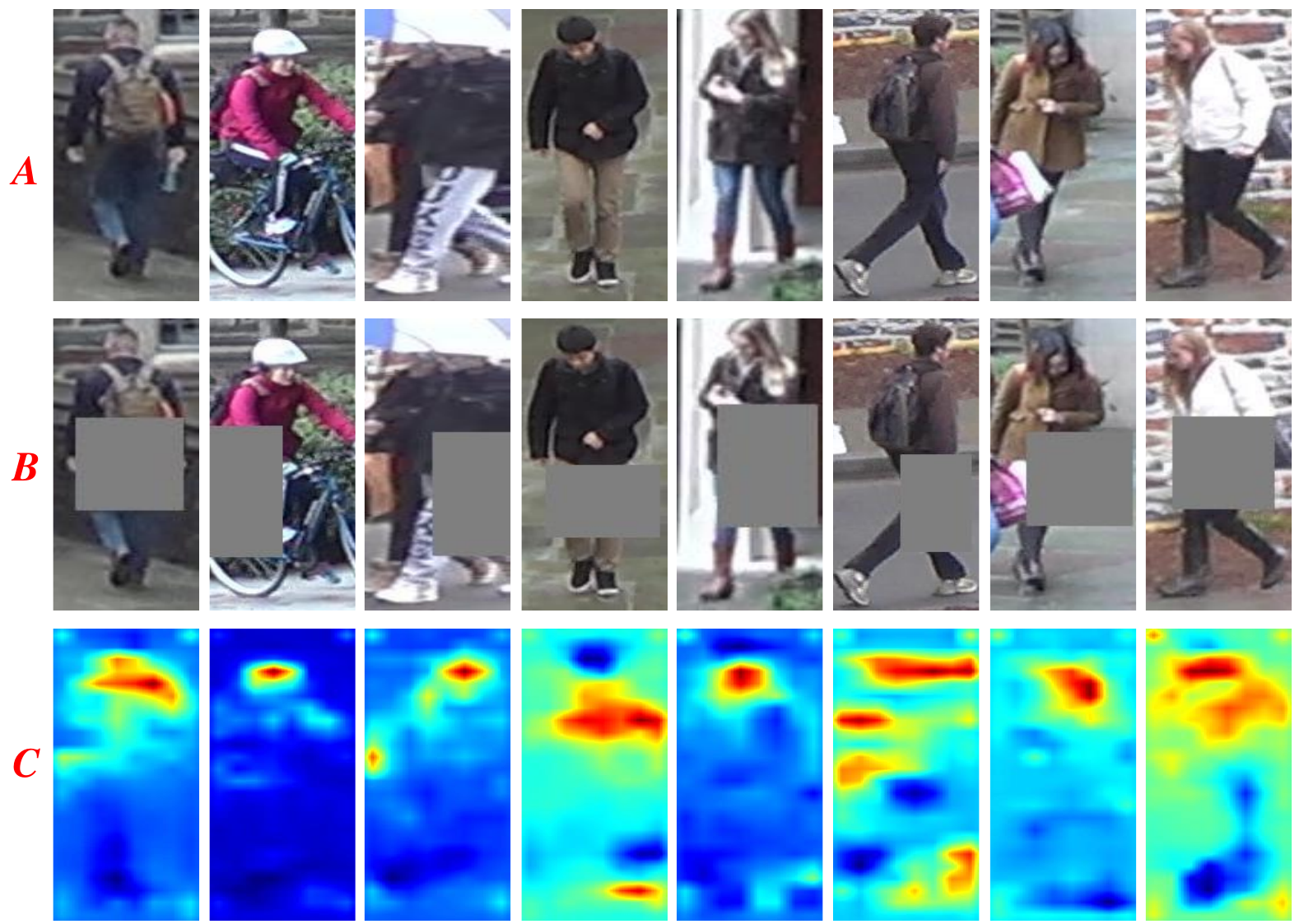}
	\end{center} 
	\caption{Visualization of feature map corresponding to random manual occlusion.  \textcolor{red}{Image Group A (the first row ) is  the normal person  images.  We randomly erase part of Image Group A to get Image Group B (the second row). The Image Group C (the  third  row) is corresponding  feature maps  of  Image Group B.}    }
	\label{Occluded_Image} 
\end{figure}

\textbf{Occluded Person Re-ID.} 
A feature of MHSA-Net is that it handles Re-ID of occluded persons well. So, we also compared MHSA-Net with a series  of occluded person re-id methods on the Occluded-DukeMTMC dataset, P-DukeMTMC-reID dataset, Partial-REID dataset, and Partial-iLIDS  dataset.

\textbf{Occluded-DukeMTMC and  P-DukeMTMC-reID.}  MHSA-Net$^{*}$  only uses the local features $p(x)^{*}$  for the  occluded Re-ID  task. 
As shown in Table~\ref{SOTA-2}, on Occluded-DukeMTMC, 
our MHSA-Net$^{*}$ achieves \textbf{59.7  Rank-1 accuracy and 44.8  mAP}, which outperforms most previous methods. 
Compared with the baseline model, the MHSA-Net$^{*}$ gains  $20.8 $ Rank-1 and $19.6 $ mAP improvement. 
As shown in Table~\ref{SOTA-4}, on P-DukeMTMC-reID, 
our MHSA-Net$^{*}$ achieves \textbf{70.7  Rank-1 accuracy and 41.1  mAP}, which outperforms all the previous methods. 
Compared with the baseline model, the MHSA-Net$^{*}$ gains  $9.7 $ Rank-1 and $14.1 $ mAP improvement.

The MHSA-Net combines both global feature $q(x)^{*}$ and local feature $p(x)^{*}$ to do  the occluded person  Re-ID task.  
The global feature $q(x)^{*}$ captures global information from the whole person image. 
Hence, it inevitably encodes contents of scene regions that occlude the person, and this leads to decreased performance. 
This can be remedied by reducing the constraints on the global feature branch. 
Specifically, if we drop the $\mathcal{L}_{CE}$ in the Global Feature Branch (GFB), the extraction of global feature becomes a simple downsampling from local features, making this global feature $q(x)^{*}$ less sensitive to occlusions. We denote the pipeline using such a design as MHSA-Net$^{\dagger}$. 
In Table~\ref{SOTA-2} and Table~\ref{SOTA-4}, the performance of MHSA-Net$^{\dagger}$ is clearly better than MHSA-Net, and only slightly worse than MHSA-Net$^*$. 
The MHSA-Net$^{\dagger}$ also achieves a competitive performance in Table~\ref{SOTA-1}.

In summary, for general database, we can use MHSA-Net. 
For datasets with certain occlusions, we can use MHSA-Net$^{\dagger}$, which best balances the global and local features. 
For datasets with severe occlusions, we can use MHSA-Net$^{*}$, where local features play more important roles. 
As  shown  in Fig.~\ref{Occluded_Image},  for the manually drawn occlusion, the MHSAM can better avoid the feature extraction of the occlusion part, and better extract the key person information of the non-occlusion part.

\textbf{Partial-REID and Partial-iLIDS.} 
The comparison of Re-ID on  these  two datasets is shown in Table~\ref{SOTA-3}. 
We also trained our model using the  Market-1501 training set. 
Our MHSA-Net$^{*}$  and  MHSA-Net$^{\dagger}$ also achieve  the best performance on both datasets. 
In both of these two data settings, compared with the baseline model, MHSA-Net$^{*}$ and  MHSA-Net$^{\dagger}$  gain a large improvement on both datasets. Like in Occluded-DukeMTMC and P-DukeMTMC-reID,  MHSA-Net$^{*}$  and  MHSA-Net$^{\dagger}$  have  better performance than MHSA-Net in Partial-REID and Partial-iLIDS datasets.

\begin{table*}[h!t]
	\centering
	\caption{Results produced by combining different components of the   MHSA-Net. 
		MHSA-Net$^{\dagger}$ indicates that we  drop  the $\mathcal{L}_{CE}$ in the training  process.     
		MHSA-Net$^{*}$=MHSAM+IHTL+FDRT+ACM denotes that we   only   use the  local   feature  $p^{*}(x)$ conducts person Re-ID task. 
		From ``Baseline''  to ``MHSA-Net$^{\dagger}$'' in this  table, the parameter $K$ is set to $8$ in   implementation.}
	
	\setlength{\tabcolsep}{3.8mm}{
		\begin{tabular}{c|c c | c c | c c | c c  }
			\hline \hline
			\multirow{2}{*}{Method}   &
			
			\multicolumn{2}{|c|}{Market-1501} &\multicolumn{2}{|c}{DukeMTMC-reID} &\multicolumn{2}{|c}{CUHK03-Detected} &\multicolumn{2}{|c}{CUHK03-Labeled}  \\
			\cline{2-9} 	
			&  Rank-1 & mAP & Rank-1 & mAP & Rank-1 & mAP & Rank-1 & mAP   \\	
			\hline     
			Baseline              &   92.0    & 78.8      &   81.0    & 62.8   &   56.3    & 53.0   &   58.6    & 55.2   \\  \hline
			Baseline+MHSAM (K=8)  &   93.0    &  80.2     &   85.2    & 70.2   &  68.1     & 64.4   &  72.1    & 69.4   \\ 
			Baseline+MHSAM+ACM (K=8)  & 93.4      & 81.9      &    86.5   & 73.1  &   69.9   & 65.8   &   72.9    & 70.1    \\  
			Baseline+MHSAM+FDRT (K=8)  &   93.6    & 82.1      &  86.3   & 72.6   &    70.0  & 65.7   &  73.4    & 69.5  \\ 
			Baseline+MHSAM+IHTL  (K=8)    &   93.5    & 82.2      &  86.4   & 72.6   &    71.2  & 67.6   &  73.6    & 70.5   \\  \hline
			Baseline+MHSAM+IHTL+ACM (K=8)  &  94.1     & 83.3    &   86.8   & 72.7   &  72.2   & 69.9   &   \textbf{75.9}   & \textbf{73.4}  \\ 
			Baseline+MHSAM+FDRT+ACM (K=8)  &  94.1   & 83.6      &  86.5    & 73.1   &   71.1   & 69.4  &   72.9   & 68.9  \\ 
			Baseline+MHSAM+IHTL+FDRT (K=8)   & 93.9      & 83.2      &   86.9   & \textbf{73.2}  &    72.2   & 69.3  &  75.0  &  72.3   \\
			MHSA-Net$^{*}$  (K=8) & 94.0      & 82.9      &  86.3   &  72.5      &    72.4   & 69.7  &  74.4  & 72.0   \\  
			MHSA-Net$^{\dagger}$ (K=8)    & 94.3      & 82.5      &   87.0   &  72.6      &    72.7   & 69.9  &  75.2  & 72.3   \\  \hline
			MHSA-Net (K=5)    & 94.3               & 83.5      &    87.1  &  73.0     &    \textbf{73.4}  & \textbf{70.2}     &  75.8    &  73.0    \\
			MHSA-Net (K=6)    & 94.2      & \textbf{84.1}      &   87.0  & 73.0      &    \textbf{73.4}  & 70.1     &  75.2    &  72.8    \\
			MHSA-Net (K=8)    & \textbf{94.6}      & 84.0      &    \textbf{87.3}    & 73.1    &     72.8  & 69.3    &  75.6    &  72.7   \\
			\hline \hline
	\end{tabular} }
	\label{TableAbTotal-1} 
\end{table*}

\subsection{Ablation Study of MHSA-Net}~\label{Ablation}

We conducted ablation studies to show effectiveness of each component in the MHSA-Net. 
We show the ablation  experiments results in Tables~\ref{TableAbTotal-1} and \ref{TableAbTotal-2}. 
By  a series  of  discussions of hyper-parameters in Subsection~\ref{AbMHSAM_sub}, we found the best hyper-parameters setting for the proposed  MHSA-Net: $\lambda_1=1e-4$ and $\lambda_2=1.0$ in Eq.~\ref{eq13}; $\lambda_3=1e-3$ in Eq.~\ref{eq15};  $\gamma=1e-3$ in Eq.~\ref{eq14}. 
In these experiments, we set $K=8$ in MHSAM, as we observed it produces stable and effective person matching results.

Through ablation studies, we have the following observations: 
(1) Each individual component effectively improves the performance of the baseline model, as shown in Table~\ref{TableAbTotal-1}. 
Compared with the baseline model, the entire MHSA-Net (K=8) achieves: $2.6$  Rank-1 and $7.2$  mAP improvement on Market-1501; $6.3$  Rank-1 and $10.3$  mAP improvement on DukeMTMC-Re-ID; $16.5$  Rank-1 and $16.3$  mAP on CUHK03-Detected; $17.0$  Rank-1 and $17.5$  mAP on CUHK03-Labeled. When $K=5$ or $6$, the change in performance is minor.
(2) We show the effectiveness of each embedding in $P^{\bot}(x) \in \mathbb{R}^{K \times 512}$, i.e. $P^{\bot}(x)_i, i=1,  2, \cdots, K$, $K=8$. 
We only use the $P^{\bot}(x)_i$ to search the target person. Here, we conduct the ablation studies on Occluded-DukeMTMC and CUHK03-Detected datasets in Table~\ref{TableAbTotal-2}. As  we can see the Table~\ref{TableAbTotal-2}, compared with  the baseline model, each  feature $P^{\bot}(x)_i$    achieves  large  improvement over  the baseline model in these  two  datasets. So, it indicates that each feature in $P^{\bot}(x)$ can better carry on both the \emph{generic} and \emph{occluded} person Re-ID tasks.

\begin{table}[]
	\centering
	\caption{Results on Occluded-DukeMTMC and  CUHK03-Detected dataset for  each embedding  in  the $P^{\bot}(x)$.  The Bold is the best result. }
	\setlength{\tabcolsep}{2mm}{
		\begin{tabular}{c|c c | c c}
			\hline \hline
			\multirow{2}{*}{Method}   &
			
			\multicolumn{2}{|c|}{Occluded-DukeMTMC} &\multicolumn{2}{|c}{CUHK03-Detected}  \\
			\cline{2-5} 	
			&  Rank-1 & mAP & Rank-1 & mAP  \\	
			\hline
			Baseline    &   38.9  &  25.6    &  56.3   & 53.0    \\ 	\hline
			$P^{\bot}(x)_1$    &   55.2    &  \textbf{40.4 }     &  68.7  & 65.0  \\
			$P^{\bot}(x)_2$     &    \textbf{55.7 } &  \textbf{40.4 }   & 68.1  & 64.4  \\
			$P^{\bot}(x)_3$     &   53.3    &  39.6      &  69.4   & 66.4 \\
			$P^{\bot}(x)_4$     &   54.9    &  40.1      &  68.6   & 64.7  \\
			$P^{\bot}(x)_5$     &   54.5    &  39.9    & 69.1   & 66.3  \\
			$P^{\bot}(x)_6$     &   54.3    &  40.2    &  69.4   & 66.4 \\
			$P^{\bot}(x)_7$     &   53.2    &  38.7     &  \textbf{70.0 } & \textbf{67.7 } \\
			$P^{\bot}(x)_8$     &   53.8    &  39.4      &  68.9   & 65.7  \\
			\hline  \hline
			
	\end{tabular} }
	\label{TableAbTotal-2} 
\end{table}

\begin{figure}\centering
	\begin{center} 
		\includegraphics[scale=0.3]{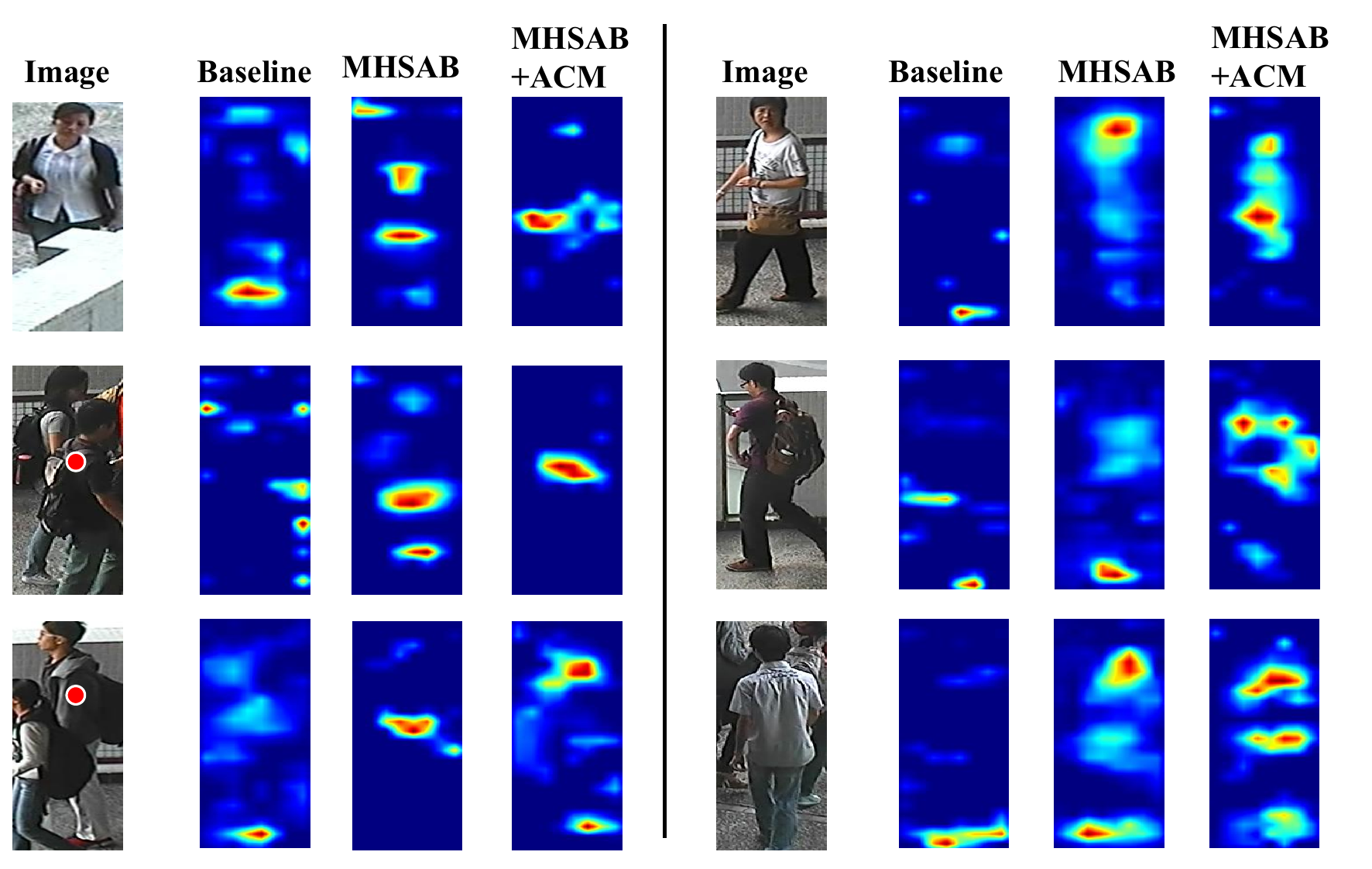}
	\end{center} 
	\caption{Visualization of attention maps from Baseline, MHSAB and MHSAB+ACM. As shown in column four and
		eight, the  attention areas from MHSAB+ACM can  effectively locate on the key suregions.} 
	\label{Fig8} 
\end{figure}

\begin{figure*}[h]
	\centering
	\begin{center} 
		\includegraphics[scale=0.18]{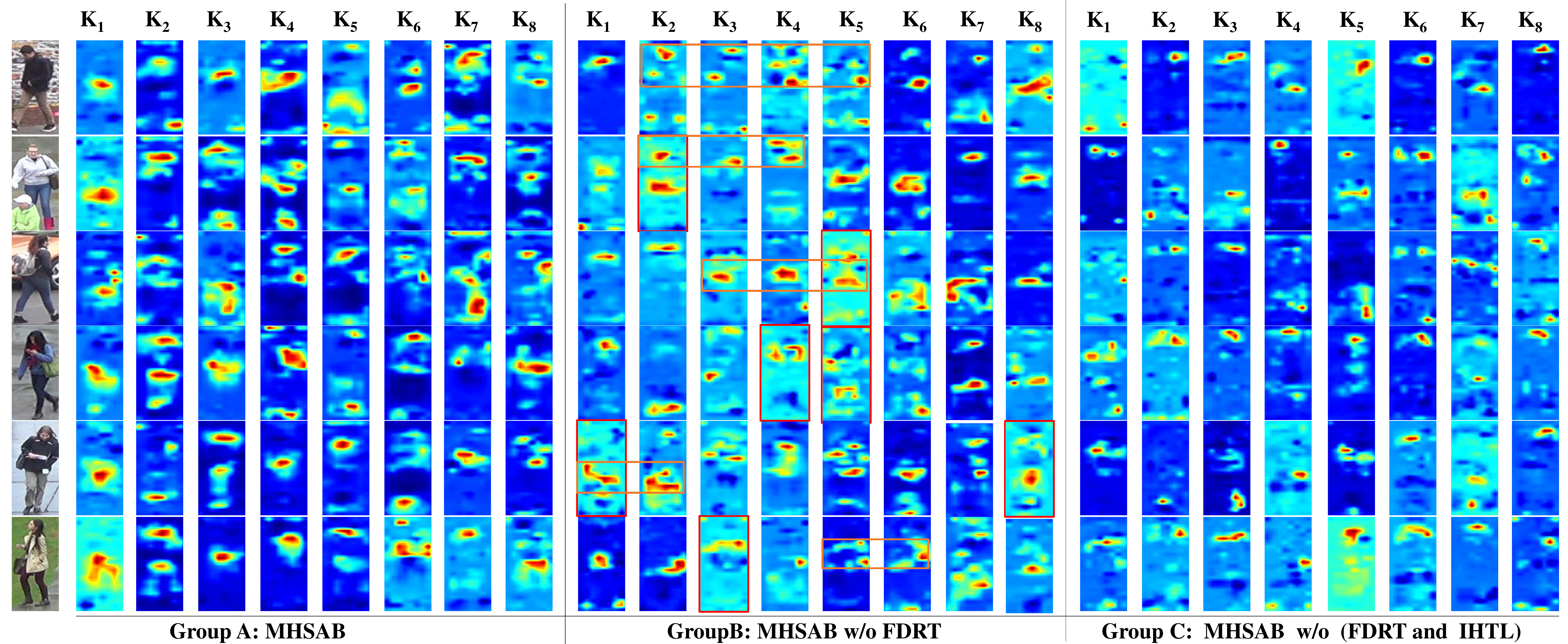}
	\end{center} 
	\caption{The feature maps of the $K=8$ feature  branches (from K$_1$ to K$_8$) of MHSAB under different variants.
			The feature maps  corresponding to $K_i, (i=1, 2, \cdots, K)$ represents the person  information captured by the $K_i$-th attention head in  MHSAB.} 
	\label{Fig_MHSAM_11} 
\end{figure*}

In the visualization results in Fig.~\ref{Fig8}, compared with Baseline, MHSAB  can effectively capture  more  key  sub-regions. 
Based on the MHSAB,  we  introduce  the ACM into MHSAB, i.e. MHSAB+ACM. Compared  with MHSAB,  some attention information is  suppressed and some attention information is highlighted. And subjectively,  we can see  that the highlight attention areas conducted  by MHSAB+ACM are more  important  than that of MHSAB.   Besides, the left half of the Fig.~\ref{Fig8},   compared with baseline model,  our  MHSAB, and  MHSAB+ACM  can   better  capture the key information  from  the unoccluded  regions in the occlusion person images.

Besides,  as  shown in  Fig.~\ref{Fig_MHSAM_11}, we visualized the feature maps of the $8$ feature  branches (from K$_1$ to K$_8$) of MHSAB under different variants.  The feature maps  corresponding to $K_i, (i=1, 2, \cdots, K)$ represents the person  information captured by the $K_i$-th attention head in  MHSAB.
(I) As  shown in ``\textbf{Group A: MHSAB}'' of  Fig.~\ref{Fig_MHSAM_11},  the Feature Diversity  Regularization Term (FDRT) can     help   the MHSAB    effectively  capture   diversity     information  for  person matching.  Since  this, MHSAM  can capture  key information from different perspectives, which enhances the model robustness.
(II) If  we  remove  the FDRT  from  MHSAB,  i.e. ``\textbf{Group B: MHSAB w/o FDRT}'', the feature maps (from K$_1$ to K$_8$)   are  mixed with some redundant information, and some feature responses (in red box especially ) are scattered and weak.  
(III) If  we  remove  the FDRT  and Improved Hard  Triplet  Loss(IHTL)  from  MHSAB,  i.e. ``\textbf{Group B: MHSAB w/o FDRT and IHTL}'', the responses of key inforamtion in the feature maps  become sparse and weak.
Without the constraint of IHTL  and FDRT, it is difficult to ensure that every feature branch  head  in MHSAB can capture the key diversity information for  person matching.

In  all,   results  in   Table~\ref{TableAbTotal-1} can  sufficiently  evidence  that  the effectiveness of each component.

\subsection{Hyper-parameters Discussion  in  MHSA-Net}~\label{AbMHSAM_sub}

\subsubsection{Multi-Head Self-Attention Mechanism (MHSAM)}~\label{AbMHSAM}

In this module, we (1) try to find suitable $K$ for MHSAM; and (2) discuss  the influence of different mechanisms in MHSAM. 

(1) Table~\ref{Tabel_AbMHSAM-1} shows  the main results in the Market-1501 and CUHK03-Detected datasets. We set $\lambda_1=0$ and $\lambda_2=0$ in Eq.~\ref{eq13}, and $\lambda_3=0$ in Eq.~\ref{eq15}. Here, we set $K=1,2,\cdots,9$ in MHSAM. 
As the results show, MHSAM with $K > 0$ can effectively improve the baseline's  performance in both datasets. 
And we find 
$K=7,8$ are suitable parameters when MHSAM produces good performance in both datasets over both measures. 

(2) Table~\ref{Tabel_AbMHSAM-2} show the influence of different mechanisms in MHSAM on the Occluded-DukeMTMC dataset. 
First, the effects of different feature fusion operations to $P^{\bot}(x) \in \mathbb{R}^{K \times 512}$ in MHSAM are compared: MHSA-Net$^{*}$ uses   SAFFM to fuse the $P^{\bot}(x)$ and produces a $512$-dimensional vector; 
MHSA-Net$^{*}$ (CONCAT) directly   concatenates the $P^{\bot}(x)$ to one  $(K+1) \times 512$ vector;  MHSA-Net$^{*}$ (SUM) simply sums up the  $P^{\bot}(x)$ to one $512$-D vector. 
The SAFFM results in the best performance in these three feature fusion operations.  The output vector from SAFFM also has much lower dimension than that from feature concatenation. 
Second, the effect of the ``Residual Learning Module'' is compared. 
``MHSA-Net$^{*}$ w/o RLM'' drops the ``Residual Learning Module'' from MHSA-Net$^{*}$, and this leads to declined performance. 
Hence, ``Residual Learning Module''  is beneficial for MHSA-Net$^{*}$  to capture useful local information. 

\begin{table}[!t]
	\centering
	\caption{Results on Market-1501 and  CUHK03-Detected testing Data from different hyper-parameter $K$ in the MHSAM.  The Bold is the best result.}
	\setlength{\tabcolsep}{2mm}{
		\begin{tabular}{c|c c | c c}
			\hline \hline
			\multirow{2}{*}{Method}   &
			
			\multicolumn{2}{|c|}{Market-1501} &\multicolumn{2}{|c}{CUHK03-Detected}  \\
			\cline{2-5} 	
			&  Rank-1 & mAP & Rank-1 & mAP  \\	
			\hline
			Baseline    &   92.0    &  78.8      &  56.3   & 53.0  \\ 	\hline
			$K=1$    &   93.0    &  80.8      &  65.5   & 61.2  \\
			$K=2$     &    93.0  &  80.8    & 66.1  & 62.0  \\
			$K=3$     &   93.1    &  80.7      &  66.1   & 62.9 \\
			$K=4$     &   93.2    &  80.5      &  \textbf{68.3 }  & 64.0  \\
			$K=5$     &   92.7    &  80.6    & 68.2   & 64.8  \\
			$K=6$     &   92.8    &  80.0    &  67.4   & 64.6  \\
			$K=7$     &   \textbf{93.3 }   &  \textbf{81.0 }    &  \textbf{68.9 }  & 65.1  \\
			$K=8$     &   93.0    &  80.2      &  68.1   & 64.4  \\
			$K=9$     &   92.6   &  80.1     &  \textbf{68.6 }  & \textbf{65.4 } \\ \hline  \hline
	\end{tabular} }
	\label{Tabel_AbMHSAM-1} 
\end{table}

\begin{table}[h!t]
	\centering
	\caption{ Influence of  different strategy  on the  Occluded-DukeMTMC dataset  in the MHSAM. Here, the  hyper-parameter $K$ in the  MHSAM is $8$. The Bold is the best result.}
	
	\setlength{\tabcolsep}{2.5mm}{
		\begin{tabular}{c| c c c  c}
			\hline \hline
			\multirow{2}{*}{Method}   &
			\multicolumn{4}{|c}{Occluded-DukeMTMC}  \\
			\cline{2-5} 	
			&   Rank-1 & Rank-5 &  Rank-10 &  mAP  \\	
			\hline
			MHSA-Net$^{*}$   &  \textbf{59.7 } &  \textbf{74.3 } & \textbf{79.5 } &  \textbf{44.8 }  \\
			MHSA-Net$^{*}$ (CONCAT) & 51.7  &  68.1  & 73.5   & 36.3   \\ 
			MHSA-Net$^{*}$ (SUM) & 53.8  &  70.9  & 76.4  & 38.8   \\ 
			MHSA-Net$^{*}$ w/o RLM &   55.9  &  72.4  & 77.4  & 41.8   \\
			\hline \hline
	\end{tabular} }
	\label{Tabel_AbMHSAM-2} 
\end{table}

\subsubsection{Feature Diversity Regularization Term (FDRT)} ~\label{AbFDRT}

\begin{figure}[h!t]
	\begin{center} 
		\includegraphics[scale=0.34]{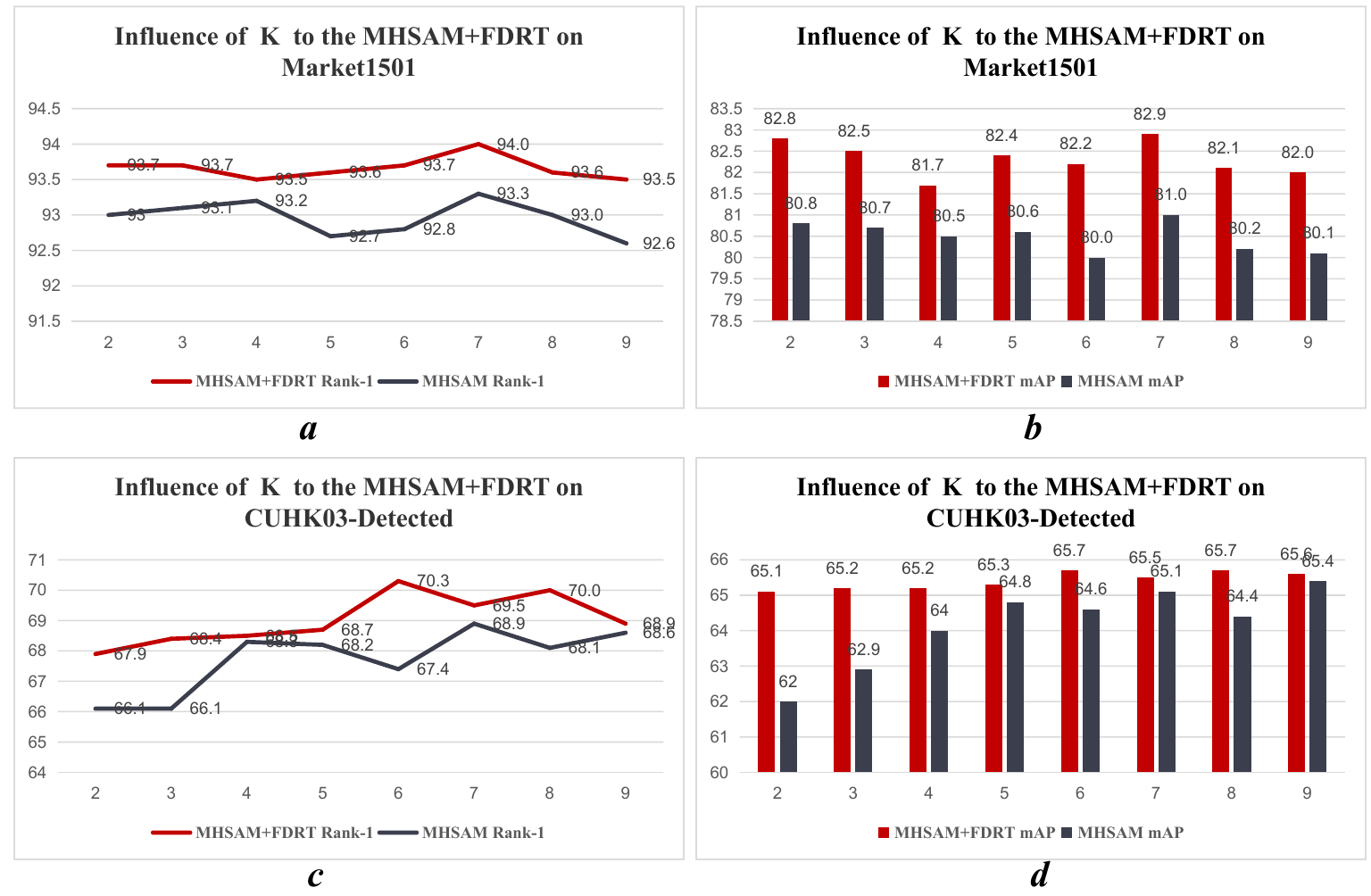}
	\end{center} 
	\caption{Results  of  MHSAM+FDRT  and MHSAM in the  Market-1501  and CUHK03-Detected testing data from different hyper-parameters $K$ in MHSAM. Here, the hyper-parameter $\lambda_1=10^{-4}$.  \textcolor{red}{Figure \textbf{a} shows the influence  of the parameter \textbf{K} on the MHSAM+FDRT  under Rank-1 score in  Market-1501 dataset. Figure \textbf{b} shows the influence  of the parameter \textbf{K} on the MHSAM+FDRT  under mAP score in  Market-1501 dataset.  Figure \textbf{c} shows the influence  of the parameter \textbf{K} on the MHSAM+FDRT  under Rank-1 score in  CUHK03-Detected dataset.  Figure \textbf{d} shows the influence  of the parameter \textbf{K} on the MHSAM+FDRT  under mAP score in  CUHK03-Detected dataset. } } 
	\label{Figure_FDRT-2} 
\end{figure}

This section discusses the suitable $\lambda_1$ for $\mathcal{L}_{FDRT}$, and FDRT's performance under different values of hyper-parameter $K$ in MHSAM. 
Table~\ref{Tabel_FDRT-1} and Fig.~\ref{Figure_FDRT-2} show  the results. 
We set $\lambda_2=0$ (Eq.~\ref{eq13}) and $\lambda_3=0$ (Eq.~\ref{eq15}). 

Table~\ref{Tabel_FDRT-1} shows Rank-1 and mAP results under different $\lambda_1$,  from $10^{-6}$ to $10^{-1}$. Both Rank-1 and mAP reaches the highest score when $\lambda_1=10^{-4}$. 

Then, with this setting of $\lambda_1=10^{-4}$ (Eq.~\ref{eq10}), we discuss performance of FDRT on different hyper-parameter $K$ values in MHSAM. 
We conducted ablation studies in Market-1501 and CUHK03-Detected datasets. 
Compared with MHSAM (in Orange polylines and bars Figs.~\ref{Figure_FDRT-2}), 
adding FDRT in MHSAM (in Blue polylines and bars, respectively) improves the performance of MHSAM in person Re-ID.

\begin{table}[!t]
	\centering
	\caption{Results on CUHK03-Detected testing data from different hyper-parameters $\lambda_1$ in Feature Diversity Regularization Term (FDRT).   The Bold is the best result. }
	
	\setlength{\tabcolsep}{6mm}{
		\begin{tabular}{c| c c}
			\hline \hline
			\multirow{2}{*}{Method}   &
			\multicolumn{2}{|c}{CUHK03-Detected}  \\
			\cline{2-3} 	
			&   Rank-1 & mAP  \\	
			\hline
			Baseline+MHSAM    &   68.1   & 64.4   \\  \hline
			$K=8, \lambda_1=10^{-6}$       &  69.6  & \textbf{65.8 } \\ 
			$K=8, \lambda_1=10^{-5} $        &  69.7   & 64.9  \\
			$K=8, \lambda_1=10^{-4} $   &  \textbf{70.0 } & 65.7  \\
			$K=8, \lambda_1=10^{-3} $  &  69.4   & 65.3  \\
			$K=8, \lambda_1=10^{-2}$   &   69.1   & 65.4  \\
			$K=8, \lambda_1=10^{-1}$    &  69.1  & 65.3  \\ \hline
			\hline
	\end{tabular} }
	\label{Tabel_FDRT-1} 
\end{table}

\begin{figure}[h!t]
	\begin{center} 
		\includegraphics[scale=0.34]{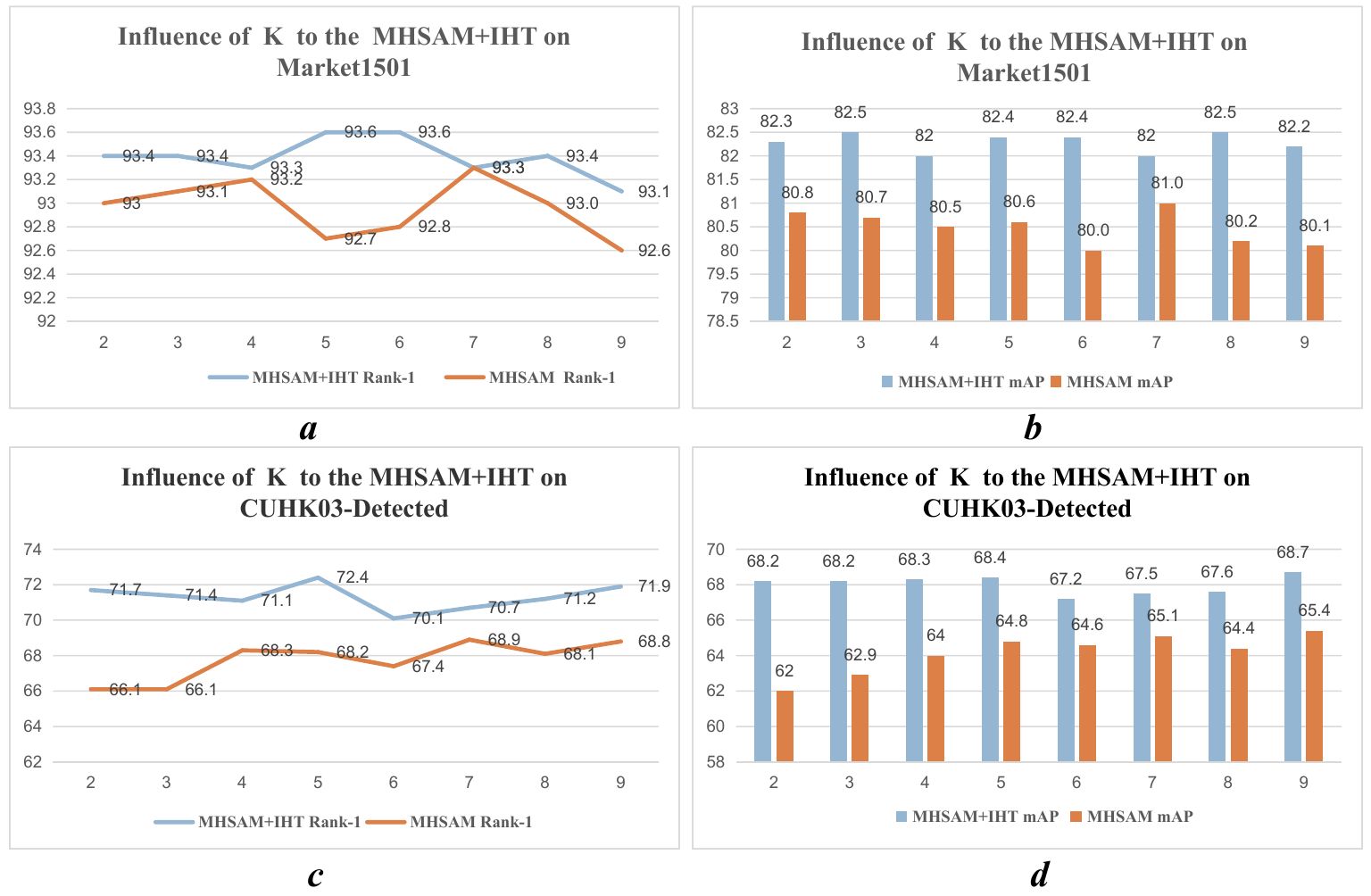}
	\end{center} 
	\caption{Results  of the   MHSAM+IHTL  and the  MHSAM in the  Market-1501  and CUHK03-Detected testing data from different hyper-parameter $K$ in MHSAM. Here, the hyper-parameter $\lambda_2=1.0$. \textcolor{red}{Figure \textbf{a} shows the influence  of the parameter \textbf{K} on the MHSAM+IHT  under Rank-1 score in  Market-1501 dataset. Figure \textbf{b} shows the influence  of the parameter \textbf{K} on the MHSAM+IHT  under mAP score in  Market-1501 dataset.  Figure \textbf{c} shows the influence  of the parameter \textbf{K} on the MHSAM+IHT  under Rank-1 score in  CUHK03-Detected dataset.   Figure \textbf{d} shows the influence  of the parameter \textbf{K} on the MHSAM+IHT  under mAP score in  CUHK03-Detected dataset.} } 
	\label{Figure_IHT-2} 
\end{figure}

\subsubsection{Improved  Hard  Triplet Loss (IHTL)} ~\label{AbIHT}  

For this module, we find the suitable weight $\lambda_2$ in IHTL, and discuss the performance of different $K$ values in MHSAM. 
Table~\ref{Tabel_IHT-1} and Fig.~\ref{Figure_IHT-2} show the main results. Here we set $\lambda_1=0$ (Eq.~\ref{eq13}) and $\lambda_3=0$ (Eq.~\ref{eq15}).  

Table~\ref{Tabel_IHT-1} shows that with the setting of $\lambda_2$ from $0.01$ to $100$, Rank-1 and mAP scores get better and then decline. 
For both $K=7$ and $8$, IHTL gets the best performance when $\lambda_2=1.0$. 

Then, with $\lambda_2=1.0$ set, we compare the performance of different $K$ in MHSAM. 
We conducted ablation  studies in Market-1501 and CUHK03-Detected datasets. 
Compared with the MHSAM (Orange polylines and bars in Fig.~\ref{Figure_IHT-2}), introducing IHTL into MHSAM,  i.e. MHSAM+IHTL (Blue polylines and bars in Fig.~\ref{Figure_IHT-2}), improves the performance of MHSAM in person Re-ID. 

\begin{table}[!t]
	\centering
	\caption{Results on CUHK03-Detected testing Data from different hyperparameters $\lambda_2$ in Improved Hard Triplet  Loss (IHTL). Here $K=7,8$ in the  MHSAM. The Bold is the best result. }
	
	\setlength{\tabcolsep}{4mm}{
		\begin{tabular}{c| c c}
			\hline \hline
			\multirow{2}{*}{Method}   &
			\multicolumn{2}{|c}{CUHK03-Detected}  \\
			\cline{2-3} 	
			&   Rank-1 & mAP  \\	
			\hline
			$K=7$ Baseline+MHSAM      &  68.9   & 65.1     \\ \hline
			$K=7, \lambda_2=10^{2} $        &  61.2  & 59.5  \\
			$K=7, \lambda_2=10^{1} $        &  67.8  & 65.5 \\
			$K=7, \lambda_2=1.0 $             &  70.7  & 67.5  \\
			$K=7, \lambda_2=10^{-1}$         &  69.8   & 65.2  \\
			$K=7, \lambda_2=10^{-2}$         &  68.6   & 64.8  \\  \hline
			$K=8$  Baseline+MHSAM          &  68.1   & 64.4   \\    \hline
			$K=8, \lambda_2=10^{2} $        &  63.4  & 61.1  \\
			$K=8, \lambda_2=10^{1} $        &  69.9  & 67.0 \\
			$K=8, \lambda_2=1.0 $             &  \textbf{71.2 } & \textbf{67.6 }  \\
			$K=8, \lambda_2=10^{-1}$         &  70.9   & 67.5  \\
			$K=8, \lambda_2=10^{-2}$         &  68.7   & 65.1  \\  \hline 	\hline
	\end{tabular} }
	\label{Tabel_IHT-1} 
\end{table}

\subsection{Attention  Competition  Mechanism (ACM)}~\label{ACM_ablation}

In this module, we try to find suitable hyper-parameters $\lambda_3$ (Eq.~\ref{eq15}) and $\gamma$ (Eq.~\ref{eq14}) in the Attention  Competition  Mechanism (ACM). 
Table~\ref{Tabel_ACM} shows the main results. 
We conducted the experiments on the CUHK03-Detected test dataset with $K=8$ set in MHSAM.

The competition loss functions in ACM  are $\mathcal{L}_{MHSAB} = \mathcal{L}_{SAFFM}+\lambda_1 \mathcal{L}_{FDRT}+ \lambda_2 \mathcal{L}_{IHTL} + \mathcal{L}_{ReN}$ and $\mathcal{L}_{C}$. 
The $\mathcal{L}_{FDRT}$ and $\mathcal{L}_{IHTL}$ are two terms we proposed here. 
To demonstrate the effectiveness of ACM individually, we set $\lambda_1=0$ and $ \lambda_2=0$ in $\mathcal{L}_{MHSAB}$.
(Table~\ref{TableAbTotal-1} in Setion~\ref{Ablation} shows the effectiveness of ACM combined with other  contributions.)

Table~\ref{Tabel_ACM} shows that when  $\gamma \leq 10^{-2}$  and  $\lambda_3 \leq 10^{-1}$, the ACM effectively improves the performance of MHSAM in the person Re-ID task.   
When $\gamma$ is too big, the performance declines. 
This is because that each element in $\alpha$ belongs to $[0, 1]$. 
If we set a too-big $\gamma$, attention weights on most regions will be suppressed by $\mathcal{L}_{C}$. 
Based on these observations, 
we set $\lambda_3=10^{-3}$ and $\gamma=10^{-3}$ in our  MHSA-Net.
In  all,  based on the suitable  parameter setting,   ACM can effectively improve  the performance  of  the MHSAB.   

\begin{table}[!t]
	\centering
	\caption{Results on CUHK03-Detected testing data from  the different hyperparameter  $\lambda_3$ and $\gamma$ in attention regularization term~\ref{eq11}. Here, $K=8$ in the  MHSAM. The Bold is the best result. }
	
	\setlength{\tabcolsep}{4mm}{
		\begin{tabular}{c| c c}
			\hline \hline
			\multirow{2}{*}{Method}   &
			 \multicolumn{2}{|c}{CUHK03-Detected }  \\
			\cline{2-3} 	
			&   Rank-1 & mAP  \\	
			\hline
			Baseline+MHSAM      &  68.1   & 64.4  \\ 	\hline
			$\lambda_3=10^{-5}, \gamma=10^{-3} $       &  68.7  & 65.1  \\
            $\lambda_3=10^{-4}, \gamma=10^{-3} $       &  69.1   & 65.0  \\
            $\lambda_3=10^{-3}, \gamma=10^{-3} $       &  \textbf{69.9 }  & 65.8  \\
            $\lambda_3=10^{-2}, \gamma=10^{-3} $       &  69.0  & 65.3  \\
            $\lambda_3=10^{-1}, \gamma=10^{-3} $       & 69.4   & 65.1  \\
            $\lambda_3=1, \gamma=10^{-3} $        &  58.4   & 55.6 \\ \hline
            $\lambda_3=10^{-3}, \gamma=10^{-4} $       &  68.8   & \textbf{65.9 } \\
            $\lambda_3=10^{-3}, \gamma=10^{-2} $       & 69.5   & 65.4  \\
            $\lambda_3=10^{-3}, \gamma=10^{-1} $       &  67.9   & 64.5  \\
            $\lambda_3=10^{-3}, \gamma=5 \times 10^{-1} $       &  67.4   & 64.2  \\
            $\lambda_3=10^{-3}, \gamma=1 $       &  66.9   & 63.8  \\
             \hline  \hline
	\end{tabular} }
	\label{Tabel_ACM} 
\end{table}

\section{Limitation  and Discussion}

Although our proposed MHSA-Net  achieved  the competitive   performance  in the occluded  person Re-ID task, some limitations  and discussion must be taken into consideration.  

Firstly, our proposed MHSA-Net  mainly considers the situation of objects occluding person. 
The situation of person  occluding  person  is not considered in the proposed MHSA-Net. 
Therefore, it is necessary to continue to optimize MHSA-NET in future  work,   to improve the model performance on   the situation of person occluding person.

Besides, the person search  requires a combination of  the  person  detection task and  the person  Re-ID task in actual scenarios.  
Only relying on the person Re-ID model  cannot effectively search for  the  target person.  
Therefore, it is necessary to study how to combine MHSAN-Net with person  detection models to build an end-to-end person  search framework

\section{Conclusion}
We proposed a Multi-Head Self-Attention Network (MHSA-Net) to improve the ability of the Re-ID model on capturing the key  information from the occluded person image. 
Specifically, we introduced the Multi-Head Self-Attention Branch (MHSAB) to adaptively capture  key local person  information, and produce multiple diversity embeddings of one person image  to facilitate person matching.  
We also designed an Attention Competition Mechanism (ACM) to  further help MHSAB prune out non-important local information.  
Extensive experiments were conducted to validate the effectiveness of each component in MHSA-Net; and they showed that MHSA-Net achieves competitive performance on three  standard  person Re-ID datasets and four occlusion person Re-ID datasets.

\section*{Acknowledgments}

This work is supported by  National Key R\&D Program of China (No. 2021ZD0111900),  National Natural Science Foundation of China (No.U21B2038, 61876012, 61976040,
62172073, 61906011), National Science Foundation of USA
(OIA-1946231, CBET-2115405), Chinese Postdoctoral Science Foundation (No. 2021M700303), Liaoning Provincial Natural
Science Foundation of China (No. 2021-MS-110).
No conflict of interest: Hongchen Tan, Xiuping Liu, Baocai Yin, and Xin Li declare that they have no conflict of interest.

\ifCLASSOPTIONcaptionsoff
\newpage
\fi

{\small
	\bibliographystyle{ieee_fullname}   
	\bibliography{mybibfile}

\begin{thebibliography}{10}\itemsep=-1pt

\bibitem{Alexander2017}
Hermans Alexander, Beyer Lucas, and Leibe. Bastian.
\newblock In defense of the triplet loss for person re-identification.
\newblock In {\em arXiv:1703.07737}, 2017.

\bibitem{He2018A}
He Anfeng, Luo Chong, Tian Xinmei, and Zeng Wenjun.
\newblock A twofold siamese network for real-time object tracking.
\newblock In {\em IEEE Conference on Computer Vision and Pattern Recognition
  (CVPR)}, 2018.

\bibitem{Ashish2017}
Vaswani Ashish, Shazeer Noam, Parmar Niki, Uszkoreit Jakob, Jones Llion,
  N~Gomez n Łukasz~Kaiser Aidan, and Polosukhin Illia.
\newblock A structured self-attentive sentence embedding.
\newblock In {\em Conference and Workshop on Neural Information Processing
  Systems}, 2017.

\bibitem{8099841}
S. Bai, X. Bai, and Q. Tian.
\newblock Scalable person re-identification on supervised smoothed manifold.
\newblock In {\em 2017 IEEE Conference on Computer Vision and Pattern
  Recognition (CVPR)}, pages 3356--3365, Los Alamitos, CA, USA, jul 2017. IEEE
  Computer Society.

\bibitem{Bryan2019}
Xia Bryan~(Ning), Gong Yuan, Zhang Yizhe, and Poellabauer Christian.
\newblock Second-order non-local attention networks for person
  re-identification.
\newblock In {\em IEEE International Conference on Computer Vision}, 2019.

\bibitem{9156982}
Xuesong Chen, Canmiao Fu, Yong Zhao, Feng Zheng, Jingkuan Song, Rongrong Ji,
  and Yi Yang.
\newblock Salience-guided cascaded suppression network for person
  re-identification.
\newblock In {\em 2020 IEEE/CVF Conference on Computer Vision and Pattern
  Recognition (CVPR)}, pages 3297--3307, 2020.

\bibitem{cheng2018eccv}
Wang Cheng, Zhang Qian, Huang Chang, Liu Wenyu, and Wang Xinggang.
\newblock Mancs: A multi-task attentional network with curriculum sampling for
  person re-identification.
\newblock In {\em Proceedings of the European Conference on Computer Vision
  (ECCV)}, 2018.

\bibitem{Chi17}
Su Chi, Li Jianing, Zhang Shiliang, Xing Junliang, Gao Wen, and Tian Qi.
\newblock Pose-driven deep convolutional model for person re-identification.
\newblock In {\em IEEE International Conference on Computer Vision}, 2017.

\bibitem{AANET19}
Tay Chiat-Pin, Roy Sharmili, and Yap Kim-Hui.
\newblock Aanet: Attribute attention network for person re-identifications.
\newblock In {\em IEEE Conference on Computer Vision and Pattern Recognition},
  2019.

\bibitem{Chunfeng}
Song Chunfeng, Huang Yan, Wanli Ouyang, and Liang Wang.
\newblock Mask-guided contrastive attention model for person re-identification.
\newblock In {\em IEEE Conference on Computer Vision and Pattern Recognition
  (CVPR)}, pages 1179--1188, 2018.

\bibitem{5479416}
Lin Chunli and Wang Kejun.
\newblock A behavior classification based on enhanced gait energy image.
\newblock In {\em 2010 International Conference on Networking and Digital
  Society}, volume~2, pages 589--592, 2010.

\bibitem{9513260}
Yongxing Dai, Jun Liu, Yan Bai, Zekun Tong, and Ling-Yu Duan.
\newblock Dual-refinement: Joint label and feature refinement for unsupervised
  domain adaptive person re-identification.
\newblock {\em IEEE Transactions on Image Processing}, 30:7815--7829, 2021.

\bibitem{Di2017Diversity}
Xie Di, Xiong Jiang, and Pu Shiliang.
\newblock All you need is beyond a good init: Exploring better solution for
  training extremely deep convolutional neural networks with orthonormality and
  modulation.
\newblock In {\em CVPR}, 2017.

\bibitem{Dongreid2014}
Jian Dong, Qiang Chen, Shen Xiaohui, Yang Jianchao, and Yan Shuicheng.
\newblock Towards unified human parsing and pose estimation.
\newblock In {\em CVPR}, 2014.

\bibitem{Dzmitry14}
Bahdanau Dzmitry, Cho Kyunghyun, and Bengio Yoshua.
\newblock Neural machine translation by jointly learning to align and
  translate.
\newblock In {\em https://arxiv.org/abs/1409.0473v2}.

\bibitem{Fangyi19}
Liu Fangyi and Zhang Lei.
\newblock View confusion feature learning for person re-identification.
\newblock In {\em IEEE International Conference on Computer Vision}, 2019.

\bibitem{Pose-guided2020}
Shang Gao, Jingya Wang, Huchuan Lu, and Zimo Liu.
\newblock Pose-guided visible part matching for occluded person reid.
\newblock In {\em CVPR}, 2020.

\bibitem{Guanan2020}
Wang Guan’an, Yang Shuo, Liu Huanyu, Wang Zhicheng, Yang Yang, Wang Shuliang,
  Yu Gang, Zhou Erjin, and Sun Jian.
\newblock High-order information matters: Learning relation and topology for
  occluded person re-identification.
\newblock In {\em CVPR}, 2020.

\bibitem{Guanshuo2018}
Wang Guanshuo, Yuan Yufeng, Chen Xiong, Li Jiwei, and Zhou Xi.
\newblock Learning discriminative features with multiple granularities for
  person re-identification.
\newblock In {\em arXiv:1804.01438}, 2018.

\bibitem{7780507}
Albert Haque, Alexandre Alahi, and Li Fei-Fei.
\newblock Recurrent attention models for depth-based person identification.
\newblock In {\em 2016 IEEE Conference on Computer Vision and Pattern
  Recognition (CVPR)}, pages 1229--1238, 2016.

\bibitem{KaimingRSE2015}
Kaiming He, Xiangyu Zhang, Shaoqing Ren, and Jian Sun.
\newblock Deep residual learning for image recognition.
\newblock In {\em CVPR}, 2016.

\bibitem{thc2019}
Tan Hongchen, Liu Xiuping, Li Xin, Zhang Yi, and Yin Baocai.
\newblock Semantics-enhanced adversarial nets for text-to-image synthesis.
\newblock In {\em ICCV}, 2019.

\bibitem{Hongyu2019Diversity}
Xu Hongyu, Wang Zhangyang, Yang Haichuan, Liu Ding, and Liu Ji.
\newblock Learning simple thresholded features with sparse support recovery.
\newblock In {\em IEEE Transactions on Circuits and Systems for Video
  Technology}, 2019.

\bibitem{2018Adversarially}
Huang Houjing, Li Dangwei, Zhang Zhang, Chen Xiaotang, and Huang Kaiqi.
\newblock Adversarially occluded samples for person re-identification.
\newblock In {\em CVPR}, 2018.

\bibitem{2020Jean-Baptiste}
Cordonnier Jean-Baptiste, Loukas Andreas, and Jaggi Martin.
\newblock On the relationship between self-attention and convolutional layers.
\newblock In {\em ICLR}, 2020.

\bibitem{Imagenet2009}
Deng Jia, Dong Wei, Socher Richard, Li Li-Jia, Li Kai, and Li Fei-Fei.
\newblock Imagenet: A large-scale hierarchical image database.
\newblock In {\em CVPR}, 2009.

\bibitem{Jianlou2018}
Si Jianlou, Zhang Honggang, Li Chun-Guang, Kuen Jason, Kong Xiangfei, Kot
  Alex~C, and Wang Gang.
\newblock Dual attention matching network for context-aware feature sequence
  based person re-identification.
\newblock In {\em arXiv preprint arXiv:1803.09937}, 2018.

\bibitem{Lu2016Knowing}
Lu Jiasen, Xiong Caiming, Parikh Devi, and Socher Richard.
\newblock Knowing when to look: Adaptive attention via a visual sentinel for
  image captioning.
\newblock In {\em IEEE Conference on Computer Vision and Pattern Recognition
  (CVPR)}, 2017.

\bibitem{Jiaxu112019}
Miao Jiaxu, Wu Yu, Liu Ping, Ding Yuhang, and Yang Yi.
\newblock Pose-guided feature alignment for occluded person re-identification.
\newblock In {\em IEEE International Conference on Computer Vision}, 2019.

\bibitem{Jiaxuan2018}
Zhuo Jiaxuan, Chen Zeyu, Lai Jianhuang, and Wang Guangcong.
\newblock Occluded person re-identification.
\newblock In {\em ICME}, 2018.

\bibitem{Jimmynormalization2016}
Ba Jimmy~Lei, Kiros Jamie~Ryan, and Hinton Geoffrey~E.
\newblock Layer normalization.
\newblock In {\em arXiv preprint arXiv:1607.06450}, 2016.

\bibitem{Jing181}
Xu Jing, Zhao Rui, Zhu Feng, Wang Huaming, and Ouyang Wanli.
\newblock Attention-aware compositional network for person re-identification.
\newblock In {\em IEEE Conference on Computer Vision and Pattern Recognition},
  2018.

\bibitem{Kalayeh2018Human}
Mahdi~M. Kalayeh, Emrah Basaran, Muhittin Gokmen, Mustafa~E. Kamasak, and
  Mubarak Shah.
\newblock Human semantic parsing for person re-identification.
\newblock In {\em IEEE Conference on Computer Vision and Pattern Recognition
  (CVPR)}, 2018.

\bibitem{DBLP01232}
M.~Esat Kalfaoglu, Sinan Kalkan, and A.~Aydin Alatan.
\newblock Late temporal modeling in 3d {CNN} architectures with {BERT} for
  action recognition.
\newblock {\em CoRR}, abs/2008.01232, 2020.

\bibitem{DBLPLCS17}
Nikolaos Karianakis, Zicheng Liu, Yinpeng Chen, and Stefano Soatto.
\newblock Person depth reid: Robust person re-identification with commodity
  depth sensors.
\newblock {\em CoRR}, abs/1705.09882, 2017.

\bibitem{2015Diederik}
Diederik~P Kingma and Jimmy Ba.
\newblock Adam: A method for stochastic optimization.
\newblock In {\em ICLR}, 2015.

\bibitem{8955791}
Jun Li, Xianglong Liu, Wenxuan Zhang, Mingyuan Zhang, Jingkuan Song, and Nicu
  Sebe.
\newblock Spatio-temporal attention networks for action recognition and
  detection.
\newblock {\em IEEE Transactions on Multimedia}, 22(11):2990--3001, 2020.

\bibitem{Zheng2015Scalable}
Zheng Liang, Shen Liyue, Tian Lu, Wang Shengjin, Wang Jingdong, and Tian Qi.
\newblock Scalable person re-identification: A benchmark.
\newblock In {\em IEEE International Conference on Computer Vision (ICCV)},
  pages 1116--1124, 2015.

\bibitem{Lingxiao2018}
He Lingxiao, Liang Jian, Li Haiqing, and Sun Zhenan.
\newblock Deep spatial feature reconstruction for partial person
  reidentification: Alignment-free approach.
\newblock In {\em CVPR}, 2018.

\bibitem{LingxiaoHE19}
He Lingxiao, Wang Yinggang, Liu Wu, Liao Xingyu, Zhao He, Sun Zhenan, and Feng
  Jiashi.
\newblock Foreground-aware pyramid reconstruction for alignment-free occluded
  person re-identification.
\newblock In {\em ICCV}, 2019.

\bibitem{Lingxiao2018xl}
He Lingxiao, Sun Zhenan, Zhu Yuhao, and Wang Yunbo.
\newblock Recognizing partial biometric patterns.
\newblock In {\em arXiv preprint arXiv:1810.07399}, 2018.

\bibitem{Munaroreid}
Matteo Munaro, Andrea Fossati, Alberto Basso, Emanuele Menegatti, and Luc~Van
  Gool.
\newblock One-shot person re-identification with a consumer depth camera.
\newblock In Shaogang Gong, Marco Cristani, Shuicheng Yan, and Chen~Change Loy,
  editors, {\em Person Re-Identification}, Advances in Computer Vision and
  Pattern Recognition, pages 161--181. Springer, 2014.

\bibitem{NING2021801}
Xin Ning, Ke Gong, Weijun Li, and Liping Zhang.
\newblock Jwsaa: Joint weak saliency and attention aware for person
  re-identification.
\newblock {\em Neurocomputing}, 453:801--811, 2021.

\bibitem{PriyaGoyal2017}
Goyal Priya, Dollar Piotr, Girshick Ross, Noord-huis Pieter, Wesolowski Lukasz,
  Kyrola Aapo, Tulloch Andrew, Jia Yangqing, and He. Kaiming.
\newblock Accurate, large minibatch sgd: Training imagenet in 1 hour.
\newblock In {\em arXiv:1706.02677}, 2017.

\bibitem{9466418}
Haocong Rao, Siqi Wang, Xiping Hu, Mingkui Tan, Yi Guo, Jun Cheng, Xinwang Liu,
  and Bin Hu.
\newblock A self-supervised gait encoding approach with locality-awareness for
  3d skeleton based person re-identification.
\newblock {\em IEEE Transactions on Pattern Analysis and Machine Intelligence},
  pages 1--1, 2021.

\bibitem{Ristani2016Performance}
Ergys Ristani, Francesco Solera, Roger Zou, Rita Cucchiara, and Carlo Tomasi.
\newblock Performance measures and a data set for multi-target, multi-camera
  tracking.
\newblock In {\em European Conference on Computer Vision (ECCV)}, 2016.

\bibitem{Ruibing19}
Hou Ruibing, Ma Bingpeng, Chang Hong, Gu Xinqian, Shan Shiguang, and Chen
  Xilin.
\newblock Interaction-and-aggregation network for person re-identification.
\newblock In {\em IEEE Conference on Computer Vision and Pattern Recognition},
  2019.

\bibitem{Discriminative_2019}
Zhou Sanping, Wang Jinjun, Meng Deyu, Liang Yudong, Gong Yihong, and Zheng
  Nanning.
\newblock Discriminative feature learning with foreground attention for person
  re-identification.
\newblock {\em {IEEE} Transactions on Image Processing}, 28(9):4671 -- 4684,
  2019.

\bibitem{Shengcai2013}
Liao Shengcai, Jain Anil~K, and Li Stan~Z.
\newblock Partial face recognition: Alignment-free approach.
\newblock In {\em TPAMI}, 2013.

\bibitem{6117504}
Sabesan Sivapalan, Daniel Chen, Simon Denman, Sridha Sridharan, and Clinton
  Fookes.
\newblock Gait energy volumes and frontal gait recognition using depth images.
\newblock In {\em 2011 International Joint Conference on Biometrics (IJCB)},
  pages 1--6, 2011.

\bibitem{Tianlong2019}
Chen Tianlong, Ding Shaojin, Xie Jingyi, Yuan Ye, Chen Wuyang, Yang Yang, Ren
  Zhou, and Wang Zhangyang.
\newblock Abd-net: Attentive but diverse person re-identification.
\newblock In {\em IEEE International Conference on Computer Vision}, 2019.

\bibitem{ZhengWang2020}
Zheng Wang, Junjun Jiang, Yang Wu, Mang Ye, Xiang Bai, and Shinichi Satoh.
\newblock Learning sparse and identity-preserved hidden attributes for person
  re-identification.
\newblock {\em IEEE Transactions on Image Processing}, 29:2013 -- 2025, 2020.

\bibitem{Weireid2014}
Li Wei, Zhao Rui, Xiao Tong, and Wang Xiaogang.
\newblock Deepreid: Deep filter pairing neural network for person
  reidentification.
\newblock In {\em IEEE Conference on Computer Vision and Pattern Recognition},
  2014.

\bibitem{li2018harmonious}
Li Wei, Zhu Xiatian, and Gong Shaogang.
\newblock Harmonious attention network for person re-identification.
\newblock In {\em IEEE Conference on Computer Vision and Pattern Recognition
  (CVPR)}, pages 2285--2294, 2018.

\bibitem{WeiShi2015}
Zheng Wei-Shi, Li Xiang, Xiang Tao, Liao Shengcai, Lai Jianhuang, and Gong
  Shaogang.
\newblock Partial person reidentification.
\newblock In {\em ICCV}, 2015.

\bibitem{Wenjie19}
Yang Wenjie, Huang Houjing, Zhang Zhang, Chen Xiaotang, Huang Kaiqi, and Zhang.
  Shu.
\newblock Towards rich feature discovery with class activation maps
  augmentation for person re-identification.
\newblock In {\em IEEE Conference on Computer Vision and Pattern Recognition},
  2019.

\bibitem{WU2022108239}
Guile Wu, Xiatian Zhu, and Shaogang Gong.
\newblock Learning hybrid ranking representation for person re-identification.
\newblock {\em Pattern Recognition}, 121:108239, 2022.

\bibitem{WU2021107424}
Wanyin Wu, Dapeng Tao, Hao Li, Zhao Yang, and Jun Cheng.
\newblock Deep features for person re-identification on metric learning.
\newblock {\em Pattern Recognition}, 110:107424, 2021.

\bibitem{Xiaobin18}
Chang Xiaobin, M~Hospedales Timothy, and Xiang Tao.
\newblock Multi-level factorisation net for person re-identification.
\newblock In {\em IEEE Conference on Computer Vision and Pattern Recognition},
  2018.

\bibitem{Xuelin18}
Qian Xuelin, Fu Yanwei, Xiang Tao, Wang Wenxuan, Qiu Jie, Wu Yang, Jiang
  Yu-Gang, and Xue Xiangyang.
\newblock Pose normalized image generation for person re-identification.
\newblock In {\em European Conference on Computer Vision}, 2018.

\bibitem{9336268}
Mang Ye, Jianbing Shen, Gaojie Lin, Tao Xiang, Ling Shao, and Steven~C.H. Hoi.
\newblock Deep learning for person re-identification: A survey and outlook.
\newblock {\em IEEE Transactions on Pattern Analysis and Machine Intelligence},
  pages 1--1, 2021.

\bibitem{Yeong-Jun16}
Cho Yeong-Jun and Yoon Kuk-Jin.
\newblock Improving person reidentification via pose-aware multi-shot matching.
\newblock In {\em IEEE Conference on Computer Vision and Pattern Recognition},
  2016.

\bibitem{Yifan2018}
Sun Yifan, Zheng Liang, Yang Yi, Tian Qi, and Wang Shengjin.
\newblock Beyond part models: Person retrieval with refined part pooling (and a
  strong convolutional baseline).
\newblock In {\em Proceedings of the European Conference on Computer Vision
  (ECCV)}, 2018.

\bibitem{Yixiao2018}
Ge Yixiao, Li Zhuowan, Zhao Haiyu, Yin Guojun, Yi Shuai, and Wang Xiaogang.
\newblock Fd-gan: Pose-guided feature distilling gan for robust person
  re-identification.
\newblock In {\em NIPS}, 2018.

\bibitem{201xf8Yumin}
Suh Yumin, Wang Jingdong, Tang Siyu, Mei Tao, and Lee Kyoung~Mu.
\newblock Part-aligned bilinear representations for person re-identification.
\newblock In {\em ECCV}, 2018.

\bibitem{9157488}
Zhizheng Zhang, Cuiling Lan, Wenjun Zeng, Xin Jin, and Zhibo Chen.
\newblock Relation-aware global attention for person re-identification.
\newblock In {\em 2020 IEEE/CVF Conference on Computer Vision and Pattern
  Recognition (CVPR)}, pages 3183--3192, 2020.

\bibitem{ZHAO2022360}
Bin Zhao, Maoguo Gong, and Xuelong Li.
\newblock Hierarchical multimodal transformer to summarize videos.
\newblock {\em Neurocomputing}, 468:360--369, 2022.

\bibitem{9399800}
Bin Zhao, Haopeng Li, Xiaoqiang Lu, and Xuelong Li.
\newblock Reconstructive sequence-graph network for video summarization.
\newblock {\em IEEE Transactions on Pattern Analysis and Machine Intelligence},
  pages 1--1, 2021.

\bibitem{8718523}
Bin Zhao, Xuelong Li, and Xiaoqiang Lu.
\newblock Cam-rnn: Co-attention model based rnn for video captioning.
\newblock {\em IEEE Transactions on Image Processing}, 28(11):5552--5565, 2019.

\bibitem{ZHAO2019161}
Cairong Zhao, Kang Chen, Zhihua Wei, Yipeng Chen, Duoqian Miao, and Wei Wang.
\newblock Multilevel triplet deep learning model for person re-identification.
\newblock {\em Pattern Recognition Letters}, 117:161--168, 2019.

\bibitem{9397375}
Cairong Zhao, Xinbi Lv, Shuguang Dou, Shanshan Zhang, Jun Wu, and Liang Wang.
\newblock Incremental generative occlusion adversarial suppression network for
  person reid.
\newblock {\em IEEE Transactions on Image Processing}, 30:4212--4224, 2021.

\bibitem{8985292}
Cairong Zhao, Xinbi Lv, Zhang Zhang, Wangmeng Zuo, Jun Wu, and Duoqian Miao.
\newblock Deep fusion feature representation learning with hard mining
  center-triplet loss for person re-identification.
\newblock {\em IEEE Transactions on Multimedia}, 22(12):3180--3195, 2020.

\bibitem{ZHAO2020107014}
Cairong Zhao, Xuekuan Wang, Wangmeng Zuo, Fumin Shen, Ling Shao, and Duoqian
  Miao.
\newblock Similarity learning with joint transfer constraints for person
  re-identification.
\newblock {\em Pattern Recognition}, 97:107014, 2020.

\bibitem{ZhedongZheng}
Zheng Zhedong, Zheng Liang, and Yang Yi.
\newblock Unlabeled samples generated by gan improve the person
  re-identification baseline in vitro.
\newblock In {\em International Conference on Computer Vision (ICCV)}, 2017.

\bibitem{Zheng_2018}
Zheng Zhedong, Zheng Liang, and Yang Yi.
\newblock Pedestrian alignment network for large-scale person
  re-identification.
\newblock {\em IEEE Transactions on Circuits and Systems for Video Technology},
  29(10):3037--3045, 2019.

\bibitem{Zhedong3Dreid}
Zhedong Zheng and Yi Yang.
\newblock Person re-identification in the 3d space.
\newblock {\em CoRR}, abs/2006.04569, 2020.

\bibitem{Kaiyang19}
Kaiyang Zhou, Yongxin Yang, Andrea Cavallaro, and Tao Xiang.
\newblock Learning generalisable omni-scale representations for person
  re-identification.
\newblock {\em IEEE Transactions on Pattern Analysis and Machine Intelligence},
  pages 1--1, 2021.

\bibitem{Zhouhan2017}
Lin Zhouhan, Feng Minwei, Nogueira dos~Santos Cicero, Yu Mo, Xiang Bing, Zhou
  Bowen, and Bengio Yoshua.
\newblock A structured self-attentive sentence embedding.
\newblock In {\em International Conference on Learning Representations}, 2017.

\bibitem{Zhong2017Re}
Zhong Zhun, Zheng Liang, Cao Donglin, and Li Shaozi.
\newblock Re-ranking person re-identification with k-reciprocal encoding.
\newblock In {\em IEEE Conference on Computer Vision and Pattern Recognition
  (CVPR)}, 2017.

\bibitem{2017Zhong}
Zhong Zhun, Zheng Liang, Kang Guoliang, Li Shaozi, and Yang Yi.
\newblock Random erasing data augmentation.
\newblock In {\em arXiv:1708.04896}, 2017.

\bibitem{Zhun2020}
Zhong Zhun, Zheng Liang, Kang Guoliang, Li Shaozi, and Yang Yi.
\newblock Random erasing data augmentation.
\newblock In {\em The National Conference on Artificial Intelligence}, 2020.

\bibitem{Zichao2016}
Yang Zichao, He Xiaodong, Gao Jianfeng, Deng Li, and Smola Alex.
\newblock Stacked attention networks for image question answering.
\newblock In {\em CVPR}, 2016.

\bibitem{Zuozhuo19}
Dai Zuozhuo, Chen Mingqiang, Gu Xiaodong, Zhu Siyu, and Tan Ping.
\newblock Batch dropblock network for person re-identification and beyond.
\newblock In {\em IEEE International Conference on Computer Vision}, 2019.

\end{thebibliography}
}

\begin{IEEEbiography}[{\includegraphics[width=1in,height=1.25in,clip,keepaspectratio]{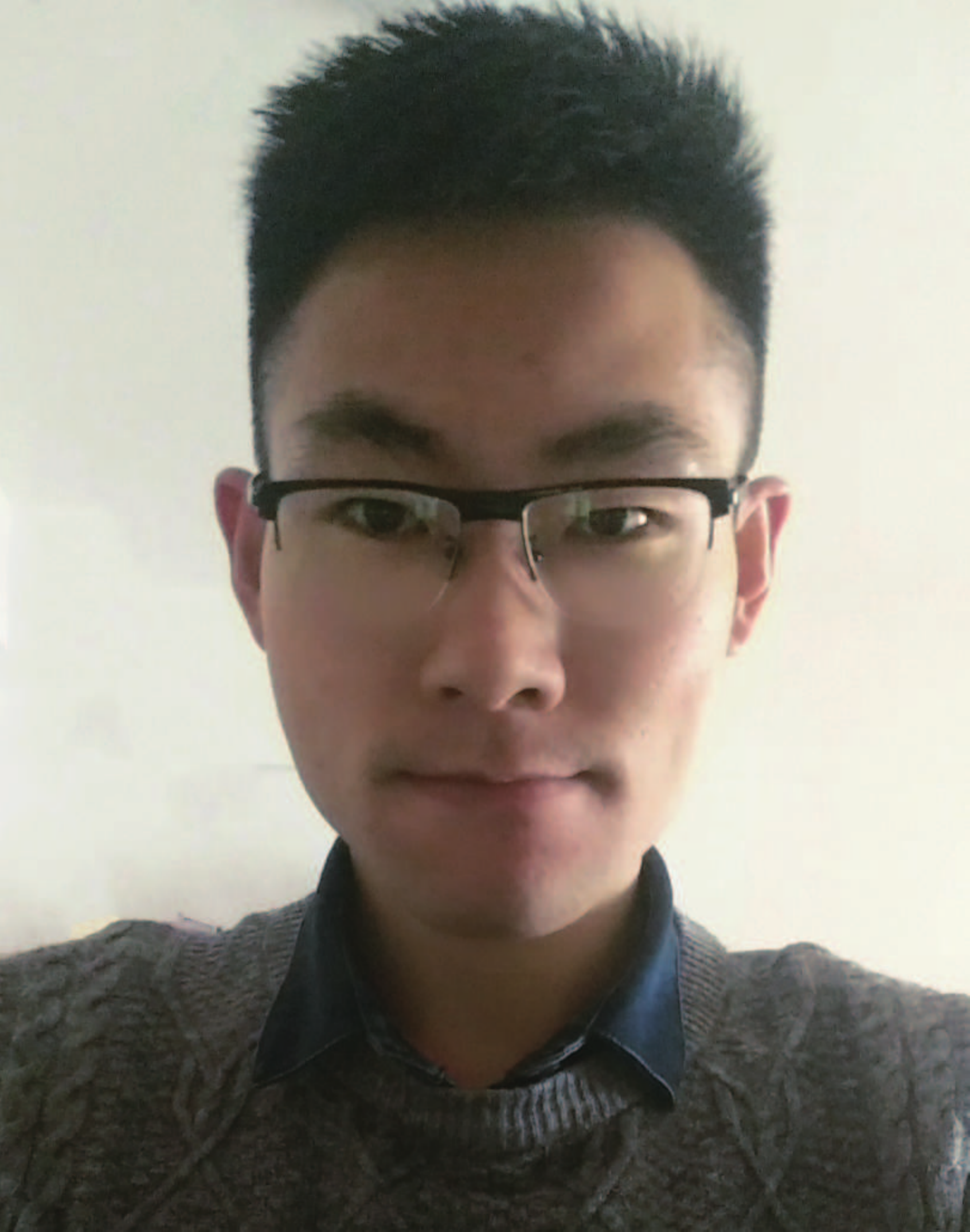}}]{Hongchen Tan} is a  Lecturer  of Artificial Intelligence Research Institute at Beijing University of Technology. He received Ph.D degrees in computational mathematics from  the Dalian University of Technology	in 2021. His research interests are  person Re-identification,  Image Synthesis, and  Object Detection.  Various parts of his work have been published in top conferences and journals, such as IEEE TIP/TMM/TCSVT/TNNLS/ICCV, and Neurocomputing. 
\end{IEEEbiography}

\begin{IEEEbiography}[{\includegraphics[width=1in,height=1.25in,clip,keepaspectratio]{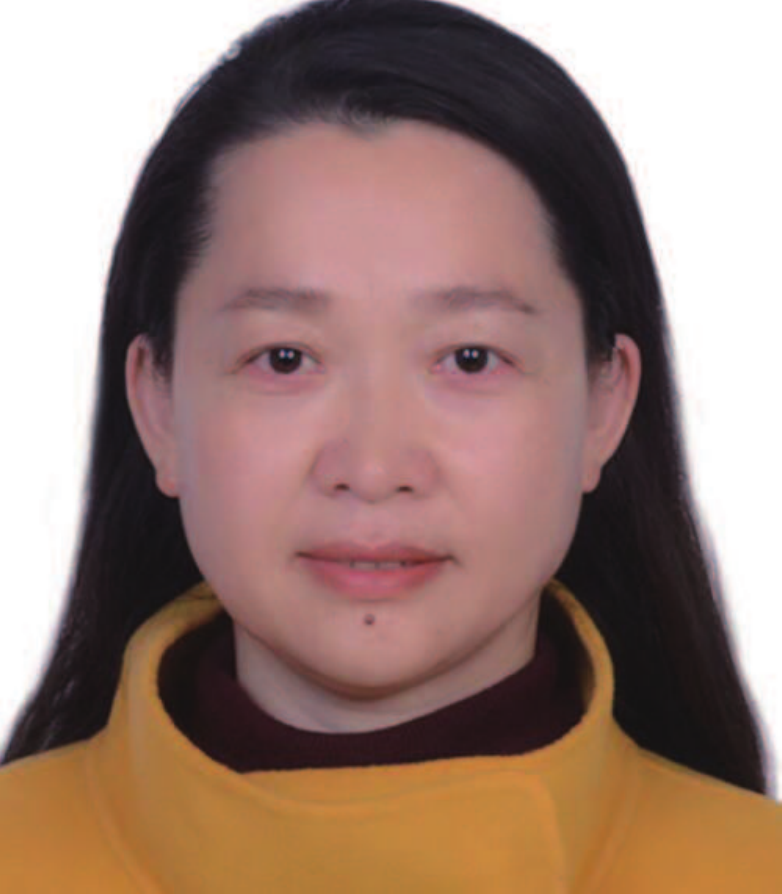}}]{Xiuping Liu}  is a Professor in School of Mathematical Sciences at Dalian University of Technology. She received Ph.D degrees in computational mathematics from Dalian University of Technology \textcolor{red}{in 1999}.   Her research interests include shape modeling and analyzing, and  computer vision.
\end{IEEEbiography}

\begin{IEEEbiography}[{\includegraphics[width=1in,height=1.25in,clip,keepaspectratio]{BaocaiYin}}]{Baocai Yin}  is  a	Professor  of Artificial Intelligence Research Institute  at  Beijing University of Technology. 	 He is	also a Researcher with the Beijing Key Laboratory of Multimedia and Intelligent Software Technology and the Beijing Advanced Innovation Center for Future Internet Technology. He received the M.S. and Ph.D. degrees in computational mathematics from the Dalian University of Technology, Dalian, China,	in 1988 and 1993, respectively.  His research interests include   multimedia, image processing, computer vision, and pattern recognition.
\end{IEEEbiography}

\begin{IEEEbiography}[{\includegraphics[width=1in,height=1.25in,clip,keepaspectratio]{XinLi}}]{Xin Li}  is a Professor at Division of Electrical \& Computer Engineering, Louisiana State University, USA. He got his B.E. degree in Computer Science from University of Science and Technology of China in 2003, and his M.S. and Ph.D. degrees in Computer Science from State University of New York at Stony Brook in 2005 and 2008.  His research interests are in Geometric and Visual Data Computing, Processing, and Understanding, Computer Vision, and Virtual Reality. For more detail, please see http://www.ece.lsu.edu/xinli.
\end{IEEEbiography}

\end{document}